\documentclass[journal=jacsat,manuscript=article]{achemso}

\usepackage[version=3]{mhchem} % Formula subscripts using \ce{}
\usepackage{caption}
\usepackage{subcaption}
\usepackage{xcolor} 
\usepackage{mathtools}
\usepackage[toc]{appendix}
\usepackage{amssymb}
\usepackage{minted}
\usepackage{algorithm, algpseudocode}
\usepackage{hyperref}

\usepackage{etoolbox} % needed for patchac
\makeatletter
\patchcmd{\acs@contact@details}{E}{*\,E}{}{}
\patchcmd{\acs@email@list@aux}{;}{\par*\,Email}{}{} % added <<<<<<<<<<<<
\makeatother

\setkeys{acs}{doi = true}

\usepackage{pdfpages}
\SectionNumbersOn
\DeclareMathOperator*{\argmax}{argmax}

\let\bs\boldsymbol

\algrenewcommand\algorithmicrequire{\textbf{Input:}}
\algrenewcommand\algorithmicensure{\textbf{Output:}}

\newcommand{\specialcell}[2][c]{%
\begin{tabular}[#1]{@{}c@{}}#2\end{tabular}}

\author{Madhav R. Muthyala}
\affiliation[uw]{Department of Chemical and Biological Engineering, University of Wisconsin--Madison, Madison, Wisconsin 53706, United States}
\affiliation[osu]
{Department of Chemical and Biomolecular Engineering, The Ohio State University, Columbus, Ohio 43210, United States}

\author{Farshud Sorourifar}
\affiliation[osu]
{Department of Chemical and Biomolecular Engineering, The Ohio State University, Columbus, Ohio 43210, United States}

\author{Tianhong Tan}
\affiliation[uw]{Department of Chemical and Biological Engineering, University of Wisconsin--Madison, Madison, Wisconsin 53706, United States}
\affiliation[osu]
{Department of Chemical and Biomolecular Engineering, The Ohio State University, Columbus, Ohio 43210, United States}

\author{You Peng}
\affiliation[dow]
{Chemometrics, AI and Statistics, Technical Expertise and Support, The Dow Chemical Company, Lake Jackson, Texas 77566, United States}

\author{Joel A. Paulson}
\email{joel.paulson@wisc.edu}
\affiliation[uw]{Department of Chemical and Biological Engineering, University of Wisconsin--Madison, Madison, Wisconsin 53706, United States}
\affiliation[osu]
{Department of Chemical and Biomolecular Engineering, The Ohio State University, Columbus, Ohio 43210, United States}

\title[]{Generative Multi-Objective Bayesian Optimization with Scalable Batch Evaluations for Sample-Efficient \textit{De Novo} Molecular Design}

\begin{document}

%\begin{tocentry}
%\includegraphics[width=\textwidth]{./figures/toc_figure_new.png}
%\end{tocentry}

\clearpage

\begin{abstract}
Designing molecules that must satisfy multiple, often conflicting objectives is a central challenge in molecular discovery. The enormous size of chemical space and the cost of high-fidelity simulations have driven the development of machine learning-guided strategies for accelerating design with limited data.
Among these, Bayesian optimization (BO) offers a principled framework for sample-efficient search, while generative models provide a mechanism to propose novel, diverse candidates beyond fixed libraries. However, existing methods that couple the two often rely on continuous latent spaces, which introduces both architectural entanglement and scalability challenges. This work introduces an alternative, modular ``generate-then-optimize'' framework for \textit{de novo} multi-objective molecular design/discovery. At each iteration, a generative model is used to construct a large, diverse pool of candidate molecules, after which a novel acquisition function, qPMHI (multi-point Probability of Maximum Hypervolume Improvement), is used to optimally select a batch of candidates most likely to induce the largest Pareto front expansion. The key insight is that qPMHI decomposes additively, enabling exact, scalable batch selection via only simple ranking of probabilities that can be easily estimated with Monte Carlo sampling. 
We benchmark the framework against state-of-the-art latent-space and discrete molecular optimization methods, demonstrating significant improvements across synthetic benchmarks and application-driven tasks. Specifically, in a case study related to sustainable energy storage, we show that our approach quickly uncovers novel, diverse, and high-performing organic (quinone-based) cathode materials for aqueous redox flow battery applications. 
\end{abstract}

\clearpage
\section{Introduction}
\label{sec:introduction}

Designing molecules that satisfy multiple competing criteria remains a grand challenge at the interface of chemistry, materials science, and artificial intelligence (AI). While high-throughput virtual screening has enabled significant advances in areas such as drug discovery \cite{lyu2019ultra, gorgulla2020open, gentile2022artificial}, catalysis \cite{greeley2006computational, yeo2021high}, and materials engineering \cite{potyrailo2011combinatorial, jain2013commentary, petousis2017high}, its brute-force nature is not well-suited to the vastness of chemical space, which is estimated to exceed $10^{60}$ synthetically accessible compounds \cite{reymond2015chemical}. In recent years, AI-guided strategies that combine predictive modeling with intelligent search have emerged as promising alternatives, offering dramatic improvements in sample efficiency and enabling discovery beyond the limitations of enumerated libraries \cite{gomez2018automatic, dave2022autonomous, merchant2023scaling, angello2024closed}.

In real-world molecular design settings, optimizing a single property in isolation is rarely sufficient. Instead, designers must navigate trade-offs among conflicting objectives. For example, organic electrode materials (OEMs) must simultaneously exhibit high redox potential for battery voltage and low solubility (in the electrolyte) to ensure cycle stability \cite{tuttle2023predicting, park2024zero}. Drug candidates must balance potency against adsorption, distribution, metabolism, and excretion (ADME) constraints \cite{tibbitts2016key}. Catalysts, in turn, must weigh activity against selectivity toward the desired products \cite{somorjai2008molecular, nabavi2024multi}. The molecular discovery literature includes a wide variety of approaches for tackling the so-called ``inverse problem'' \cite{sanchez2018inverse, pollice2021data, alshehri2022deep} of mapping properties to structure. Although many studies describe themselves as ``multi-objective,'' the distinction from single-objective or constrained optimization is often muddied. As noted in a recent review \cite{fromer2023computer}, it is common to use scalarization methods that collapse objectives into a single composite score (e.g., a weighted sum of objectives \cite{hartenfeller2008concept, ooi2018integration, fu2021mimosa, sv2022multi}), yielding only one solution per run. Rigorous multi-objective optimization (MOO) \cite{gunantara2018review, deb2016multi}, on the other hand, aims to identify the full Pareto front (i.e., the set of non-dominated solutions that collectively reveal the balance of trade-offs among objectives), thus enabling more informed downstream decision-making.

Bayesian optimization (BO) \cite{shahriari2015taking, frazier2018tutorial, paulson2024bayesian} has become a widely used framework for sample-efficient MOO, particularly when each evaluation is costly. In the conventional library-based workflow, a finite set of candidate molecules is assembled using combinatorial enumeration, retrosynthesis reconstruction, or scaffold-based heuristics. A surrogate model, such as a Gaussian process (GP) \cite{williams2006gaussian, deringer2021gaussian, griffiths2023gauche} or Bayesian neural network (BNN) \cite{ryu2019bayesian, jospin2022hands}, is trained on a modest number of labeled examples. Acquisition functions that balance predicted performance and uncertainty (such as expected hypervolume improvement) are then used to iteratively select new candidates for evaluation \cite{daulton2020differentiable}. This approach has been successfully applied to optimize the properties of organic molecules \cite{fromer2024pareto}, transition-metal complexes \cite{janet2020accurate}, and metal-organic frameworks \cite{comlek2023rapid}. However, it is inherently constrained by the need for either a fixed, fully enumerable candidate set or a continuous design space -- conditions that rarely hold in \textit{de novo} molecular discovery \cite{fromer2023computer}.

To overcome these limitations, recent work has explored integrating generative models with BO. Generative methods, which have proven highly effective in domains such as image and language modeling \cite{kumar2023comprehensive, hagos2024recent}, can be adapted to propose novel chemical structures automatically. A common approach is to train a variational autoencoder (VAE) \cite{kingma2019introduction} to map discrete molecular representations (such as SMILES strings \cite{weininger1988smiles} or molecular graphs \cite{randic1992representation}) into a continuous latent space, conduct BO in this space, and then decode the resulting latent vectors back into molecular structures. While theoretically appealing, this ``latent-optimize-then-decode'' paradigm presents several practical challenges. Decoder outputs are often invalid or highly redundant \cite{kusner2017grammar}. Latent spaces can remain high-dimensional and poorly structured, which complicates surrogate modeling and uncertainty quantification (UQ) \cite{maus2022local}. Batch acquisition, which is crucial for leveraging parallel evaluations, is typically handled using heuristics (such as the Kriging believer \cite{ginsbourger2010kriging} or local penalization \cite{gonzalez2016batch}) that have known limitations \cite{riegler2025rethinking}. Finally, the need to jointly train the encoder, decoder, and property predictor creates a high degree of architectural coupling, making the overall model fragile and difficult to train/tune.

\begin{figure*}[tb!]
    \centering
    \includegraphics[width=0.98\linewidth]{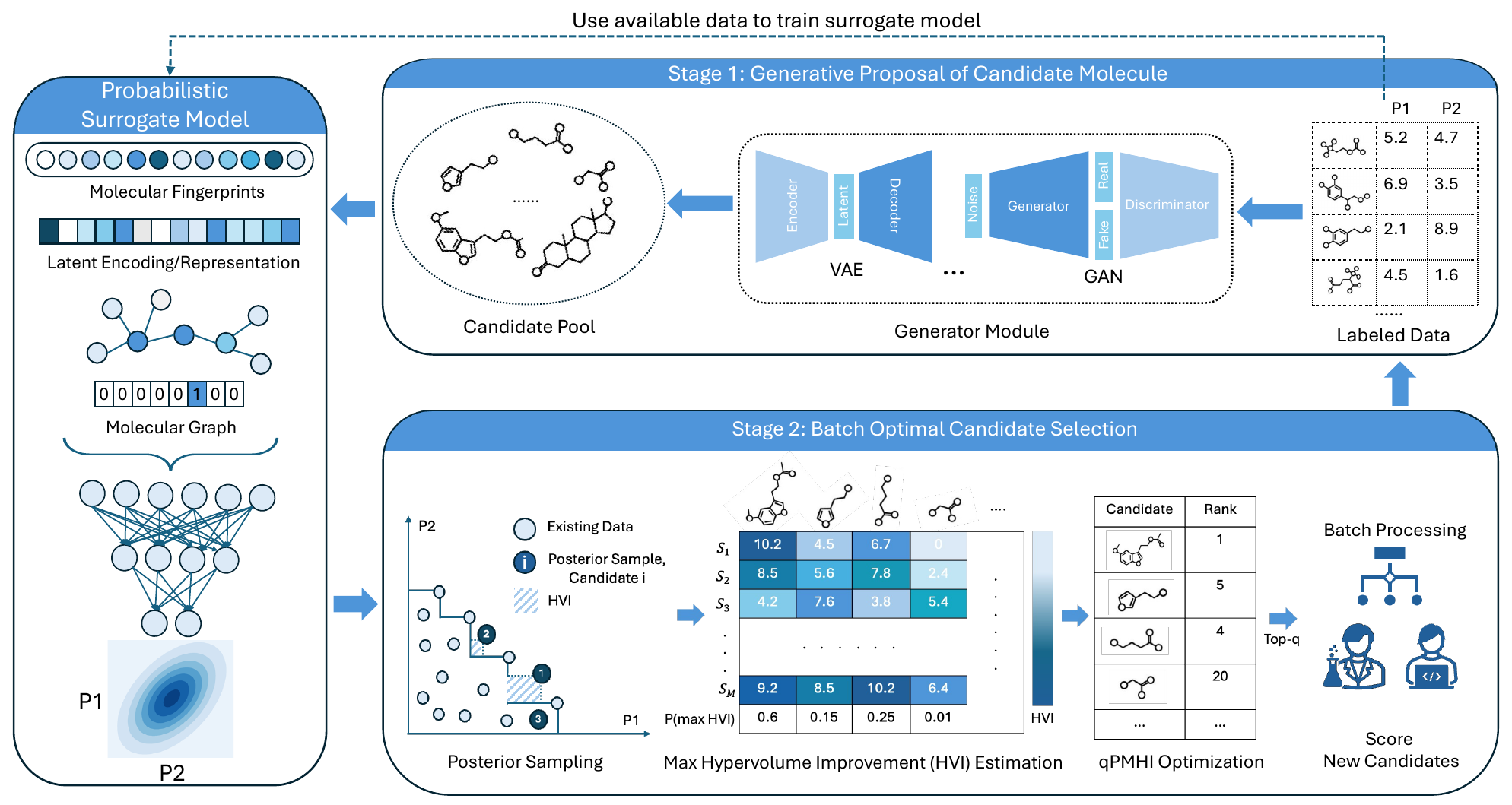}
    \caption{Overview of the proposed two-stage \textit{de novo} multi-objective molecular optimization framework. In \textbf{Stage 1}, a generative approach proposes a large pool of candidate molecules, which may be represented in diverse formats (e.g., graphs, SMILES, fingerprints). These candidates can be tailored to reflect user-defined biases toward specific regions of chemical space, high predicted property values, or high epistemic uncertainty. In \textbf{Stage 2}, a probabilistic surrogate model (trained on labeled molecules) is used to predict objective values and uncertainty. Monte Carlo posterior sampling is used to estimate the probability that each candidate achieves the maximum hypervolume improvement (HVI), yielding a simple acquisition function, qPMHI, that decomposes additively across candidates. Thus, the optimal batch can be easily selected by sorting the candidates by their maximum HVI score.}
    \label{fig:qPMHI-framework}
\end{figure*}

This work introduces a complementary methodology that reverses this paradigm: generate first, then optimize second. As illustrated in Figure \ref{fig:qPMHI-framework}, any generative model, such as a VAE \cite{kingma2019introduction}, diffusion model \cite{cao2024survey}, genetic algorithm \cite{wu2004primer}, or reinforcement learning policy \cite{franceschelli2024reinforcement}, is first used to ``dream'' a large and diverse pool of molecular candidates. This generation step can incorporate user-specified preferences (e.g., prioritizing underexplored regions or skewed property targets) and does not require uniqueness or diversity filtering. At this stage, the goal is simply to propose as many plausibly useful candidates as possible. In the second stage, a new acquisition function, referred to as \textit{qPMHI} (\textit{multi-point Probability of Maximum Hypervolume Improvement}), is used to select a batch of molecules that are most likely to induce the largest expansion of the Pareto front once queried. Because qPMHI decomposes additively across candidates, it can be optimized by selecting the top candidates ranked by a scalar acquisition score, extending ideas from the recently proposed qPO method \cite{fromer2025batched}. This additive structure allows efficient batch selection from large candidate pools without requiring any combinatorial optimization.
Our two-stage view is similar in spirit to recent work that links a policy-gradient generator with an active learning loop to oversample each epoch and then select a subset for expensive oracle calls \cite{dodds2024sample}. In contrast, we make this separation explicit and \textit{generator-agnostic}: Stage 1 can be any generator (or combination thereof), while Stage 2 performs a new and specific type of batch acquisition that supports single- and multi-objective problems as well as small to large batches.

A key strength of our proposed ``generate-then-optimize'' framework lies in its modularity. In particular, the generative model and surrogate property model can be fully decoupled, allowing for any combination of architectures or molecular representations at each stage. For example, generative decoders that are prone to producing invalid structures can be paired with encoder-only graph neural networks (GNNs) for property prediction \cite{wieder2020compact}. This decoupling also enables flexible UQ, ranging from exact GPs to deep ensembles and BNNs. By reducing architectural entanglement, the framework facilitates more robust training and easier tuning compared to the aforementioned latent-space approaches.

We validate the proposed framework through two complementary case studies. The first benchmarks performance on a widely used dataset of drug-like molecules, enabling direct comparison to several existing generative optimization methods in both single- and multi-objective settings. The second case study targets a practically relevant molecular design problem in energy storage: the discovery of quinone-based OEMs for aqueous redox flow battery applications. Here, we aim to jointly maximize redox potential (to increase cell voltage) and minimize aqueous solubility (to enhance useful cycle life). 
In both settings, our method consistently identifies high-performing candidates that expand the Pareto front with fewer queries than baseline approaches -- highlighting its broad applicability and potential effectiveness for accelerating discovery in chemically realistic design spaces.

\section{Related Work}
\label{sec:related-work}

Property-driven molecular design seeks to discover chemical structures that satisfy one or more desired properties, without relying on brute-force enumeration of all possibilities. Existing strategies generally fall into two categories: \textit{selection-based methods}, which search over a fixed library of candidate molecules, and \textit{generation-based methods}, which construct novel molecules during the search process -- potentially producing structures that have not been synthesized or even considered before. Across both categories, BO has become a leading framework for efficient discovery when each evaluation is expensive. However, the integration of BO with discrete chemical representations varies substantially depending on the design space and modeling strategy. Below, we summarize some of the main contributions across these different approaches.

\subsection{Library-based multi-objective BO for molecular design}

Some of the earliest successes in multi-objective BO for molecular design relied on large, pre-constructed libraries. These libraries were often generated using domain heuristics or combinatorial enumeration. Janet et al.~\cite{janet2020accurate} optimized redox potential and aqueous solubility over a library of 2.8 million transition-metal complexes using a neural network surrogate model with latent-distance UQ and greedy expected hypervolume improvement. Their approach achieved nearly 500-fold greater sample efficiency compared to a naive random search strategy. Similarly, Comlek et al.~\cite{comlek2023rapid} used a latent variable GP to search over a library of nearly 48,000 metal–organic frameworks (MOFs), optimizing for CO$_2$ working capacity and CO$_2$/N$_2$ selectivity using an expected maximin improvement acquisition strategy. 
MolPAL~\cite{graff2021accelerating} extends these ideas by offering a flexible active learning platform for library-based BO. It supports various multi-objective acquisition strategies and offers diversity-based pruning in both property and design spaces, including batch selection via top-$k$ ranking, $k$-means clustering in property space, or dissimilarity-based clustering in design space. However, like many such platforms, its batch selection is heuristic. More broadly, library-based BO remains fundamentally constrained by the scope of the candidate pool. While effective when the library is tractable in memory, this class of methods cannot propose truly novel structures and often misses promising regions of chemical space not captured in the initial enumeration.

\subsection{Generative methods coupled with BO}

\paragraph{Latent-optimize-then-decode frameworks.} A common approach for coupling generative modeling with BO involves embedding discrete structures (such as SMILES strings or graphs for molecules) into a continuous latent space. BO is then performed in this latent space and the optimized latent vectors are decoded back into the original discrete structures for evaluation. Gómez-Bombarelli et al.~\cite{gomez2018automatic} pioneered this strategy using a VAE trained on SMILES strings, combined with GP-based BO in the learned latent space. Grammar VAE~\cite{kusner2017grammar} extended this idea by enforcing syntactic constraints during decoding, thereby improving the rate of chemically valid molecules. More recently, LOLBO~\cite{maus2022local} introduced a local BO framework that adaptively refines the latent space around promising regions as additional data are collected. MLPS~\cite{liu2025multi} takes a different approach by learning a parametric approximation to the entire latent-space Pareto front prior to decoding. 

Although these methods are conceptually appealing, they face several practical limitations. Decoded molecules are frequently invalid or redundant, and latent spaces are often high-dimensional and poorly structured, complicating surrogate modeling and UQ. These issues are further exacerbated in the multi-objective setting, where the latent space must meaningfully represent multiple distinct properties. Batch selection is typically performed using greedy or heuristic strategies since optimizing over large batches of latent vectors can be computationally demanding and/or challenging from an optimization point of view. 

\paragraph{Acquisition-guided generation in discrete spaces.} Other methods bypass latent embeddings altogether and instead optimize acquisition functions directly over discrete chemical representations. For example, BOSS \cite{moss2020boss} uses grammar-constrained SMILES strings as input, combines them with string kernels in a GP surrogate, and employs an evolutionary algorithm to search for high-acquisition candidates. COMBO \cite{oh2019combinatorial} models the design space as a Cartesian product of categorical and ordinal variables, defines a Laplacian-based kernel over this structure, and uses GP-based BO to optimize over the resulting graph. Amortized Bayesian optimization \cite{swersky2020amortized} learns a policy network that can quickly propose high-quality candidates, effectively amortizing the acquisition maximization step. GFlowNet-based methods \cite{bengio2023gflownet} learn stochastic generation policies that sample molecules in proportion to their acquisition-based reward. Extensions to multi-objective settings \cite{zhu2023sample} have also been recently proposed.

These methods offer several advantages, including the ability to directly optimize in discrete space and more easily avoid invalid decodings. However, they still face challenges in maintaining chemical plausibility, navigating complex and highly non-convex search spaces, and scaling effectively to large batch sizes.

\subsection{Generative optimization without BO}

A broad range of molecular design methods have been developed outside the BO framework. Genetic algorithms (GAs) are a longstanding example. GB-GA \cite{brown2004graph} evolves molecular graphs using mutation and crossover while preserving chemical validity. JANUS \cite{nigam2022parallel} combines a parallel-tempered GA with a deep neural network for active learning, and demonstrates strong performance across logP and docking benchmarks. 

Reinforcement learning (RL) methods offer another powerful paradigm for exploring molecular design spaces. MolDQN \cite{zhou2019optimization} frames molecular editing as a Markov decision process and applies Q-learning to optimize scalarized multi-objective rewards. GraphAF \cite{shi2020graphaf} combines autoregressive flow models with policy gradients to generate molecules with optimized properties. PGFS \cite{gottipati2020learning} focuses on synthesizable molecules by navigating reaction spaces directly, while REINVENT~\cite{olivecrona2017molecular} uses recurrent neural networks trained on SMILES to generate candidates through RL. Recent work by S.V. et al.~\cite{sv2022multi} adapts AlphaZero to molecular optimization using Monte Carlo tree search and a fast, machine learning-derived surrogate objective, showing relatively strong performance even in multi-objective settings.

Although these methods are capable of generating diverse and chemically valid structures, they often require a large number of evaluations per run, making them less practical when querying high-fidelity simulations or experiments. This lack of sample efficiency is a primary reason we focus instead on BO-based frameworks that explicitly model epistemic uncertainty (i.e., the lack of knowledge on the structure-to-property mapping) to try and make the most out of each evaluation.

\subsection{Our contributions}

This work introduces a ``generate-then-optimize'' framework for \textit{de novo} multi-objective molecular design that decouples the generation of candidate structures from the process of selecting which ones to evaluate. In contrast to latent-space BO methods, which require tight coupling between encoders, decoders, and the surrogate models for property prediction, our approach maintains modularity between components. This flexibility allows the use of any combination of generative model and probabilistic property predictor, and enables principled acquisition-based selection over arbitrary molecular representations.
By decoupling generation from optimization, the proposed framework takes a major step toward the development of scalable, general-purpose tools for sample-efficient discovery in open-ended chemical spaces.
Our main contributions can be summarized as follows:
\begin{itemize}
    \item We generalize the idea of using an RL-based generator followed by a surrogate-based active learning loop \cite{dodds2024sample} into a modular (generator-agnostic) framework, with a novel acquisition function built for large, discrete candidate pools.
    \item We introduce a new batch acquisition function, called qPMHI (multi-point Probability of Maximum Hypervolume Improvement), that estimates the likelihood of each candidate maximally expanding the Pareto front. Because this objective decomposes additively across candidates (similarly to qPO \cite{fromer2025batched}), selecting the top-$k$ acquisition scores yields the exact optimal batch, avoiding the need for combinatorial optimization.
    \item We develop a scalable implementation of qPMHI that supports selection from large candidate pools (tens of thousands of molecules or more) on a single GPU.
    \item We conduct single- and multi-objective benchmarks comparing the proposed framework to existing (state-of-the-art) BO-based and generative optimization methods.
    \item We demonstrate the practical utility of our method through a realistic case study in sustainable energy storage, where we accelerate the discovery of high-performance quinone-based OEMs for aqueous redox flow battery applications. 
\end{itemize}

\section{Proposed Methodology}
\label{sec:proposed-method}

\subsection{Problem formulation}

We consider the problem of multi-objective optimization (MOO) over molecular structures:
\begin{align} \label{eq:moo-problem}
    \max_{\bs{x} \in \mathcal{X}} ~ \bs{f}(\bs{x}) = ( f^{(1)}(\bs{x}), \ldots, f^{(M)}(\bs{x}) ),
\end{align}
where $\bs{x}$ denotes a molecule drawn from the molecular design space $\mathcal{X}$ and $\bs{f} : \mathcal{X} \to \mathbb{R}^M$ is a vector-valued objective function composed of $M$ scalar properties. Each $f^{(m)} : \mathcal{X} \to \mathbb{R}$ represents a distinct target property that, without loss of generality, we wish to maximize. We assume all objectives are black-box and expensive to evaluate. While we focus on the noise-free case for simplicity, extensions to noisy evaluations are straightforward.

We adopt the standard Pareto dominance criterion: a molecule $\bs{x}$ dominates another $\bs{x}'$ if it performs at least as well in all objectives and strictly better in at least one. That is, $\bs{f}(\bs{x}) \succ \bs{f}(\bs{x}')$ if and only if $f^{(m)}(\bs{x}) \geq f^{(m)}(\bs{x}')$ for all $m \in \{1, \ldots, M\}$ and there exists $m'$ such that $f^{(m')}(\bs{x}) > f^{(m')}(\bs{x}')$. Completely solving \eqref{eq:moo-problem} would require us to identify the exact Pareto-optimal objective set and corresponding molecules:
\begin{subequations}
\begin{align}
    \mathcal{P}^\star &= \{ \bs{f}(\bs{x}) : \not\exists \bs{x}' \in \mathcal{X} \text{ such that } \bs{f}(\bs{x}') \succ \bs{f}(\bs{x}) \}, \\
    \mathcal{X}^\star &= \{ \bs{x} \in \mathcal{X} : \bs{f}(\bs{x}) \in \mathcal{P}^\star \}.
\end{align}
\end{subequations}
This Pareto frontier provides the trade-off surface from which a decision-maker can select a solution that adequately balances competing objectives based on downstream constraints and/or preferences. %(typically depend strongly on the specific application of interest)
In principle, the search space $\mathcal{X}$ represents all valid molecules (e.g., satisfying chemical valence rules, synthesizability constraints, etc.); however, in the \textit{de novo} setting, this space is too vast to enumerate explicitly and its size is often unknown or even unbounded. Thus, $\mathcal{X}$ should be viewed as a theoretical construct; our practical goal is not to search it exhaustively, but to identify a diverse set of high-performing candidates that meaningfully improve upon the existing (currently known) Pareto front.

Because direct evaluation of $\bs{f}$ is costly and $\mathcal{X}$ cannot be exhaustively searched, we adopt a surrogate-based sequential learning strategy, namely BO. BO constructs a probabilistic model (typically a GP, but any surrogate that predicts a distribution over outcomes can be used) to approximate the posterior $\mathbb{P}(\bs{f} \mid \mathcal{D})$ over objective values where $\mathcal{D}$ denotes the current set of observations. An acquisition function $\alpha$ is then optimized to select a batch of $q$ new candidates that are expected to be informative:
\begin{align} \label{eq:batch-selection}
\mathcal{X}_\text{acq} = \argmax_{\mathcal{X}_\text{cand} \subset \mathcal{X}, |\mathcal{X}_\text{cand}|=q} \alpha(\mathcal{X}_\text{cand}).
\end{align}
After querying $\bs{f}$ at all selected $\bs{x} \in \mathcal{X}_\text{acq}$, the dataset is updated and the surrogate model is retrained (either from scratch or fine-tuned). This loop continues until the evaluation budget is exhausted or convergence is reached.

Batch selection is particularly important in modern molecular discovery workflows, where parallel resources are often available. High-throughput simulations, multi-core computing, and experimental setups (like 96-well plates) all benefit from simultaneously evaluating multiple candidates. However, choosing an informative and diverse batch remains one of the most computationally challenging steps in BO, especially in the multi-objective setting. Evaluating the batch acquisition function $\alpha$ is already non-trivial: for example, multi-point Expected Hypervolume Improvement (qEHVI) \cite{daulton2020differentiable} involves computing expectations over the joint surrogate posterior across the batch. While gradient-based approximations exist and can be effective in continuous spaces with smooth, differentiable inputs, they are more costly than their single-objective counterparts and rely on local optimization techniques that can lead to suboptimal or redundant selections.
These challenges are magnified in discrete molecular spaces, where gradients are unavailable and the candidate space lacks a convenient parametric form. One workaround is to embed molecules into a continuous latent space using, e.g., a VAE and perform BO in this latent space. However, this introduces new difficulties, including poorly structured latent regions, low decoder validity, and the need for tight coupling between the encoder, decoder, and property predictor.

To bypass these limitations, we adopt a two-stage alternative, involving an initial generation step followed by an optimization (or selection) step, described in the next two sections. 

\subsection{Stage 1: Generative proposal of candidate molecules}

Our goal is \emph{generative} optimization: at each iteration we construct a fresh, finite pool $\widetilde{\mathcal{X}}\subset\mathcal{X}$ from which the Stage 2 selector chooses an evaluation batch. The framework is generator-agnostic (and composable), i.e., it can interface with \textit{any} type of molecular generator or combinations of generators. See the Supporting Information (Section S1) for a more detailed discussion of generative methods. By rebuilding $\widetilde{\mathcal{X}}$ at each iteration, we are continuously building novel molecules with targeted exploration around promising scaffolds. In our experiments, we instantiate Stage 1 with several options, including a lightweight surrogate-guided genetic algorithm operating on SMILES strings; implementation details appear in Section \ref{sec:results} and the Supporting Information (Section S2.2).

The size of $\widetilde{\mathcal{X}}$ is a tunable parameter. The main idea in this work is that we can easily create on the order of $10^4$ to $10^6$ molecules using existing (relatively inexpensive) generators. However, we cannot evaluate the true objective function $\bs{f}$, which requires a high-fidelity simulation or wet-lab experiment, across the full $\widetilde{\mathcal{X}}$ ever (much less at every iteration). Thus, Stage 2 of our framework performs this down-selection to a practical batch size $q$ (typically between tens to hundreds) for evaluation. The most effective pool would be the one that best balances \textit{exploitation} of current high-performers and \textit{exploration} of the unknown parts of the chemical space. Enforcing this balance in the generator itself can be quite nontrivial, as it is inherently challenging to deal with the discrete, multi-objective nature of the problem. By decoupling generation (diverse, valid proposal) from selection (acquisition-driven subset choice), we argue the generator can play to its strengths while Stage 2 handles the combinatorial decision-making.

Note that ensuring validity and synthesizability of the generated molecules is important in practice but outside the scope of this paper. We treat Stage 1 as a plug-in interface that can incorporate validity-enforcing representations, synthesizability-aware generators, and/or post-hoc filters as they mature (e.g., projecting candidates into synthesizable chemical space \cite{luo2024projecting}). Future advances in this area can be adopted with no change to Stage 2, effectively upgrading the proposal mechanism while retaining the multi-objective BO-based selector. Our results suggest that, given a fixed generator, our two-stage framework can consistently improve selection quality and, in turn, identification of higher-quality candidates.

\subsection{Stage 2: Batch optimal candidate selection}

The hypervolume of a finite approximation of the Pareto set $\mathcal{P}$ is defined as
\begin{align}
    \text{HV}( \mathcal{P}; \bs{r} ) = \lambda_M \left( \bigcup_{ \bs{y} \in \mathcal{P} } [\bs{r}, \bs{y}] \right),
\end{align}
where $\bs{r} \in \mathbb{R}^{M}$ is a dominated reference point (chosen such that every incumbent objective vector on the Pareto front dominates it), $[\bs{r},\bs{y}]$ is the hyper-rectangle spanned by $\bs{r}$ and $\bs{y}$, and $\lambda_M$ denotes the $M$-dimensional Lebesgue measure. For a new candidate $\bs{x}$ with unknown objectives $\bs{f}(\bs{x})$, its hypervolume improvement (HVI) relative to the current front $\mathcal{P}$ is
\begin{align} \label{eq:hvi}
    \Delta \text{HV}( \bs{x} ) = \text{HV}( \mathcal{P} \cup \{ \bs{f}( \bs{x} ) \}; \bs{r} ) - \text{HV}( \mathcal{P}; \bs{r} ),
\end{align}
where we omit the dependence on $\mathcal{P}$ and $\bs{r}$ for simplicity. 
Under the posterior $\mathbb{P}(\bs{f} \mid \mathcal{D})$, $\Delta \text{HV}( \bs{x} )$ is a random variable. Our proposed (batch-level) acquisition function, qPMHI (multi-point Probability of Maximum Hypervolume Improvement), is then defined as
\begin{align} \label{eq:qPMHI-acq}
    \alpha_\text{qPMHI}( \mathcal{X}_\text{cand} ) = \text{Pr}\left( \argmax_{\bs{x}' \in \widetilde{\mathcal{X}}} \Delta \text{HV}( \bs{x}' ) \in \mathcal{X}_\text{cand}  \mid \mathcal{D} \right),
\end{align}
where $\mathcal{X}_\text{cand} = \{ \bs{x}_i \}_{i=1}^q$ is a subset of $q$ candidates drawn from the current generated pool $\widetilde{\mathcal{X}}$. Under the mild assumption that the surrogate posteriors are continuous with small but nonzero noise, the probability of ties in HVI values is zero, and thus the events $\{ \bs{x} = \argmax_{\bs{x}' \in \widetilde{\mathcal{X}}} \Delta \text{HV}( \bs{x}' ) \}_{\bs{x} \in \widetilde{\mathcal{X}}}$ are mutually exclusive. This allows the acquisition score to be decomposed into a simple sum over individual candidate probabilities, i.e., 
\begin{align}
    \alpha_\text{qPMHI}( \mathcal{X}_\text{cand} ) &= \sum_{ \bs{x} \in \mathcal{X}_\text{cand} } p(\bs{x}), \quad \text{where} \quad p(\bs{x}) = \text{Pr}\left( \bs{x} = \argmax_{\bs{x}' \in \widetilde{\mathcal{X}}} \Delta \text{HV}( \bs{x}' ) \mid \mathcal{D} \right).
\end{align}
This additive structure enables batch selection via simple ranking: we compute $p(\bs{x})$ for each candidate in $\widetilde{\mathcal{X}}$ and choose the top-$q$ with highest probability. As a result, increasing the batch size $q$ incurs no combinatorial overhead. Note that $\sum_{ \bs{x} \in \widetilde{\mathcal{X}} } p(\bs{x}) = 1$, since the events form a mutually exclusive and exhaustive partition of outcomes.

It is useful to compare qPMHI to one of the most popular MOO acquisition functions qEHVI, defined as $\alpha_\text{qEHVI}( \mathcal{X}_\text{cand} ) = \mathbb{E}[ \Delta \text{HV}( \mathcal{X}_\text{cand} ) \mid \mathcal{D} ]$. Because qEHVI depends on the joint distribution of $\bs{f}( \mathcal{X}_\text{cand} )$, optimizing it requires searching over all $q$-sized subsets, which is especially difficult in discrete domains (as the size grows exponentially with $q$). By contrast, qPMHI avoids this bottleneck by reducing the acquisition evaluation to independent candidate-wise probabilities.

To estimate $p(\bs{x})$ in practice, we use Monte Carlo sampling. Depending on the surrogate model, this can be done in one of two ways. For parametric models (such as BNNs), we sample a realization of the model parameters and evaluate the resulting function on all candidates $\bs{x} \in \widetilde{\mathcal{X}}$. For GPs, we typically draw from the joint posterior over the candidate set by computing the posterior mean and covariance and sampling from the resulting multivariate normal. An alternative and more efficient method, described by Wilson et al. \cite{wilson2020efficiently}, uses a hybrid function-space and weight-space representation via random Fourier features, and applies to any covariance function with a known spectral decomposition.
In either case, we draw $L$ independent samples from the posterior to obtain $ \{ \bs{f}^{(\ell)}( \bs{x} ) \}_{\bs{x} \in \widetilde{\mathcal{X}}}$ for all $\ell = 1, \ldots, L$.
For each sample $\ell$, we (i) compute $\Delta \text{HV}^{(\ell)}( \bs{x} )$ for every candidate $\bs{x} \in \widetilde{\mathcal{X}}$, and (ii) determine the unique maximizer $\bs{x}^{\star (\ell)} = \argmax_{ \bs{x} \in \widetilde{\mathcal{X}} } \Delta \text{HV}^{(\ell)}( \bs{x} )$. The Monte Carlo estimate of $p(\bs{x})$ is:
\begin{align}
    \widehat{p}(\bs{x}) = \frac{1}{L} \sum_{ \ell = 1 }^L \mathbf{1}\{ \bs{x}^{\star (\ell)} = \bs{x} \},
\end{align}
where $\mathbf{1}\{\cdot\}$ is the indicator function that equals 1 when its argument is true and 0 otherwise. Since each posterior draw is independent, this computation can be trivially parallelized. The final acquisition $\alpha_\text{qPMHI}(\mathcal{X}_\text{cand}) \approx \sum_{i=1}^q \widehat{p}(\bs{x}_i)$ is then maximized exactly by sorting, making batch selection scalable and efficient.

We can interpret qPMHI as an extension of the recent qPO acquisition \cite{fromer2025batched} to the MOO setting. In fact, qPMHI simplifies to qPO when we replace the HVI metric with one that measures improvement in the single-objective optimal value over the discrete candidate set. 

\subsection{Workflow integration and iterative loop}

Algorithm \ref{alg:qphmi} summarizes the complete generative multi-objective batch BO procedure using the proposed qPMHI acquisition function. Each iteration begins by updating the probabilistic surrogate model using all available labeled property data (Line 2). The generator may optionally be updated using this same data (Line 3), which is important for ensuring the ``on-the-fly'' generated pool of candidates $\widetilde{\mathcal{X}}_t$ is well aligned with the multi-objective learning task. Stage 1 (the generative proposal step) is executed in Line 4, while Stage 2 (the optimization and selection step) is carried out in Lines 5 to 13. Note that the function $\text{top}$-$k(\{ \bs{x}_i, y_i \}_{i=1}^N, k)$ in Line 13 sorts the $N$ input-output pairs by their scalar output values in descending order and returns the top $k$ associated inputs. This procedure is trivially parallelizable and scales well with both batch size $q$ and pool size $N$.
The final step (Line 14) involves querying the expensive oracle $\bs{f}$ to evaluate the selected candidates and augmenting the dataset with these newly acquired labels/measurements. This iterative process continues until a fixed number of iterations $T$ is completed, or until another user-defined stopping criterion is met (easily incorporated at the end of each iteration). 

\begin{algorithm}[htb!]
\caption{Generative multi-objective batch Bayesian optimization using qPMHI}
\begin{algorithmic}[1]
\Require Generator $p_\text{gen}$, pool size $N$, batch size $q$, multi-objective oracle function $\bs{f}$, initial data $\mathcal{D}_0$, number of Monte Carlo samples $L$, number of iterations $T$, reference point $\bs{r}$.
\For{$t = 1, 2, \ldots, T$}
\State Train posterior surrogate model $\mathbb{P}( \bs{f} \mid \mathcal{D}_{t-1} )$ using available data $\mathcal{D}_{t-1}$.
\State (Optional) Update generator using available data $p_{\text{gen}}( \bs{x}) \leftarrow p_\text{gen}( \bs{x} \mid \mathcal{D}_{t-1} )$.
\State Generate pool $\widetilde{\mathcal{X}}_t = \{ \widetilde{\bs{x}}_1, \ldots , \widetilde{\bs{x}}_N \}$ via $\widetilde{\bs{x}}_{t, i} \sim p_{\text{gen}}( \bs{x} )$ for all $i=1,\ldots,N$.
\For{$\ell = 1, 2, \ldots, L$}
\State Sample function from the posterior $\bs{f}_{t}^{(\ell)} \sim \mathbb{P}( \bs{f} \mid \mathcal{D}_{t-1} )$.
\State Evaluate HVI on candidate pool $\Delta\text{HV}_{t}^{(\ell)}( \bs{x} )$ for all $\bs{x} \in \widetilde{\mathcal{X}}_t$ using $\bs{f}_{t}^{(\ell)}$.
\State Compute unique HVI maximizer $\bs{x}^{\star (\ell)}_t = \argmax_{ \bs{x} \in \widetilde{\mathcal{X}}_t } \Delta \text{HV}^{(\ell)}_t( \bs{x} )$.
\EndFor
\For{$i = 1, 2, \ldots, N$}
\State Compute probability of maximum HVI score $\widehat{p}_{t, i} = \frac{1}{L} \sum_{\ell=1}^L \mathbf{1}\{ \bs{x}^{\star (\ell)}_t = \widetilde{\bs{x}}_i \}$.
\EndFor
\State Select optimal batch $\mathcal{X}_\text{acq} \leftarrow \text{top}$-$k( \{ \widetilde{\bs{x}}_{t, i}, \widehat{p}_{t, i} \}_{i=1}^N, q )$.
\State Evaluate oracle for all $\bs{x} \in \mathcal{X}_\text{acq}$ and update data $\mathcal{D}_t \leftarrow \mathcal{D}_{t-1} \cup \{ (\bs{x}, \bs{f}( \bs{x} )) \}_{ \bs{x} \in \mathcal{X}_\text{acq} }$.
\EndFor
\Ensure Acquired data $\mathcal{D}_T$ and final posterior surrogate model $\mathbb{P}( \bs{f} \mid \mathcal{D}_T )$.
\end{algorithmic}
\label{alg:qphmi}
\end{algorithm}

A key practical consideration is that fewer than $q$ candidates may have nonzero probability of achieving the maximum HVI. In such cases, instead of directly taking the top-$q$ candidates in Line 13, we recommend a simple fallback: populate the remainder of the batch using the candidates with the highest probability of lying on the Pareto front. Although this strategy lacks the additive decomposability property that enables qPMHI’s efficient ranking-based optimization procedure, it is easy to implement and we found it to provide relatively strong empirical performance. Studying more principled alternatives -- especially those with theoretical guarantees -- is an interesting direction for future research.

Finally, domain-specific constraints can be readily incorporated into the workflow. For known constraints (such as bounds on molecular weight or violations of chemical valency), filtering can be applied during generation by rejecting infeasible candidates in Stage 1. For black-box constraints on unknown properties (e.g., toxicity or solubility), a constrained variant of qPMHI can be used in Stage 2. This involves modifying the maximization step in \eqref{eq:qPMHI-acq} to restrict attention to candidates predicted to satisfy all constraints under the surrogate model. Because qPMHI relies on sample-based approximations, such constraints can be enforced on a per-sample basis within each Monte Carlo draw. Importantly, this does not affect the overall structure of the acquisition function, i.e., the resulting constrained variant still satisfies the additive decomposability property that enables efficient batch selection.

\section{Results and Discussion}
\label{sec:results}

In this section, we evaluate the effectiveness of our proposed framework for generative multi-objective batch BO on two case studies. The first is a modified version of a widely used benchmark in molecular optimization focused on drug-like molecule discovery. The second highlights the applicability of our approach to a practical energy storage problem: the design of OEMs for aqueous redox flow batteries. We target simultaneous optimization of redox potential and aqueous solubility (two properties linked to device performance and long-term stability) using state-of-the-art property prediction models trained on domain-specific data.

The code and data used in the experiments, including an efficient implementation of Algorithm~\ref{alg:qphmi}, are available at: \href{https://github.com/PaulsonLab/Generative_MOBO_qPMHI}{https://github.com/PaulsonLab/Generative\_MOBO\_qPMHI}. Note that a more detailed description of how the methods were implemented and some additional results/analyses are provided in the Supporting Information (Sections S2 to S4). 

\subsection{Multi-objective optimization benchmark}
\label{subsec:multi-objective}

\subsubsection{Problem description}

We begin with a widely used molecular design benchmark involving the simultaneous maximization of the water–octanol partition coefficient (logP) and minimization of topological polar surface area (TPSA). This setup is adapted from Gómez-Bombarelli et al.~\cite{gomez2018automatic}, with molecules drawn from ZINC-250k, a curated subset of the ZINC database~\cite{irwin2012zinc}. 
Higher logP values are generally associated with improved membrane permeability, while lower TPSA values typically reduce hydrogen bonding capacity and promote passive diffusion, particularly across the blood-brain barrier. Because these objectives often conflict in drug design, they serve as a canonical test case for MOO methods~\cite{wang2023molecular}. 

To mitigate known pathologies when optimizing logP in an unconstrained setting, we enforce two mild constraints: SMILES string length $\leq 108$ and synthetic accessibility (SAScore)~\cite{ertl2009estimation} $\leq 8$. Note that these constraints are not intended to reflect realistic drug discovery criteria, but rather to match assumptions in prior benchmarking work. Both logP and TPSA are computed using \texttt{RDKit}~\cite{rdkit}.

\subsubsection{Baseline methods}

We evaluate four representative baselines spanning different molecular optimization strategies: (i) \textbf{VAE+BO}~\cite{gomez2018automatic}, a latent optimize-then-decode framework using a VAE generator and qEHVI acquisition; (ii) \textbf{JANUS}~\cite{nigam2022parallel}, a parallel-tempered genetic algorithm guided by an internal property predictor; (iii) \textbf{Graph-GA}~\cite{jensen2019graph}, a GA-based method operating over molecular graphs; and (iv) \textbf{MolDQN}~\cite{zhou2019optimization}, a deep reinforcement learning method that applies Q-learning to molecular graph edits.

JANUS, Graph-GA, and MolDQN are single-objective by default, so we optimize a composite objective given by the equally weighted sum of min-max normalized logP and TPSA values (based on the initial data). Notably, Graph-GA ranked second overall in a recent benchmark study by Gao et al.~\cite{gao2022sample}. The top-ranked method, REINVENT~\cite{olivecrona2017molecular}, relies on a recurrent neural network (RNN) trained on SMILES from the ChEMBL database~\cite{gaulton_chembl}; since this differs from our ZINC-250k-based setup, we evaluate REINVENT separately in Section~S4 of the Supporting Information.

\subsubsection{Implementation details and setup}

Each method is run for $T = 20$ iterations with a batch size of $q = 50$. The initial training set $\mathcal{D}_0$ consists of 2000 randomly sampled molecules from ZINC-250k. All methods are evaluated over five independent trials, with matched random seeds for fairness.

Our approach uses $L = 256$ Monte Carlo samples to estimate acquisition scores, and a candidate pool size of $N = 5000$ per iteration. The surrogate model is a custom Bayesian graph neural network (BGNN) adapted from Ramani and Karmakar~\cite{ramani2024graph}; we implemented it in \texttt{torchbnn} with post-training variance calibration following Rasmussen et al.~\cite{rasmussen2023uncertain}. 
For candidate generation, we adopt a GA inspired by the STONED method~\cite{nigam2021beyond}, which leverages the surrogate to attempt to balance exploration and exploitation. We denote this as \textbf{Ours (GA/BGNN)}. To highlight the importance of generator choice, we also include a variant \textbf{Ours (VAE/GP)} that mirrors the VAE+BO setup within our framework (i.e., uses the exact same VAE generator and GP surrogate).

Additional implementation details, including hyperparameter settings, are provided in Section~S2 of the Supporting Information. For all baselines, we preserve default or recommended settings from the original open-source repositories. While the hyperparameter values are expected to impact results, exhaustive hyperparameter tuning is impractical and beyond the scope of this study. We also note that the per-iteration computational cost for all methods in this benchmark are summarized in Section S2.3 of the Supporting Information; they are on the order of minutes for all methods, which are negligible when compared to realistic settings where oracles are expensive simulations or experiments.

\subsubsection{Pareto optimization results}

\begin{figure}[tb]
\centering
\includegraphics[width=0.8\textwidth]{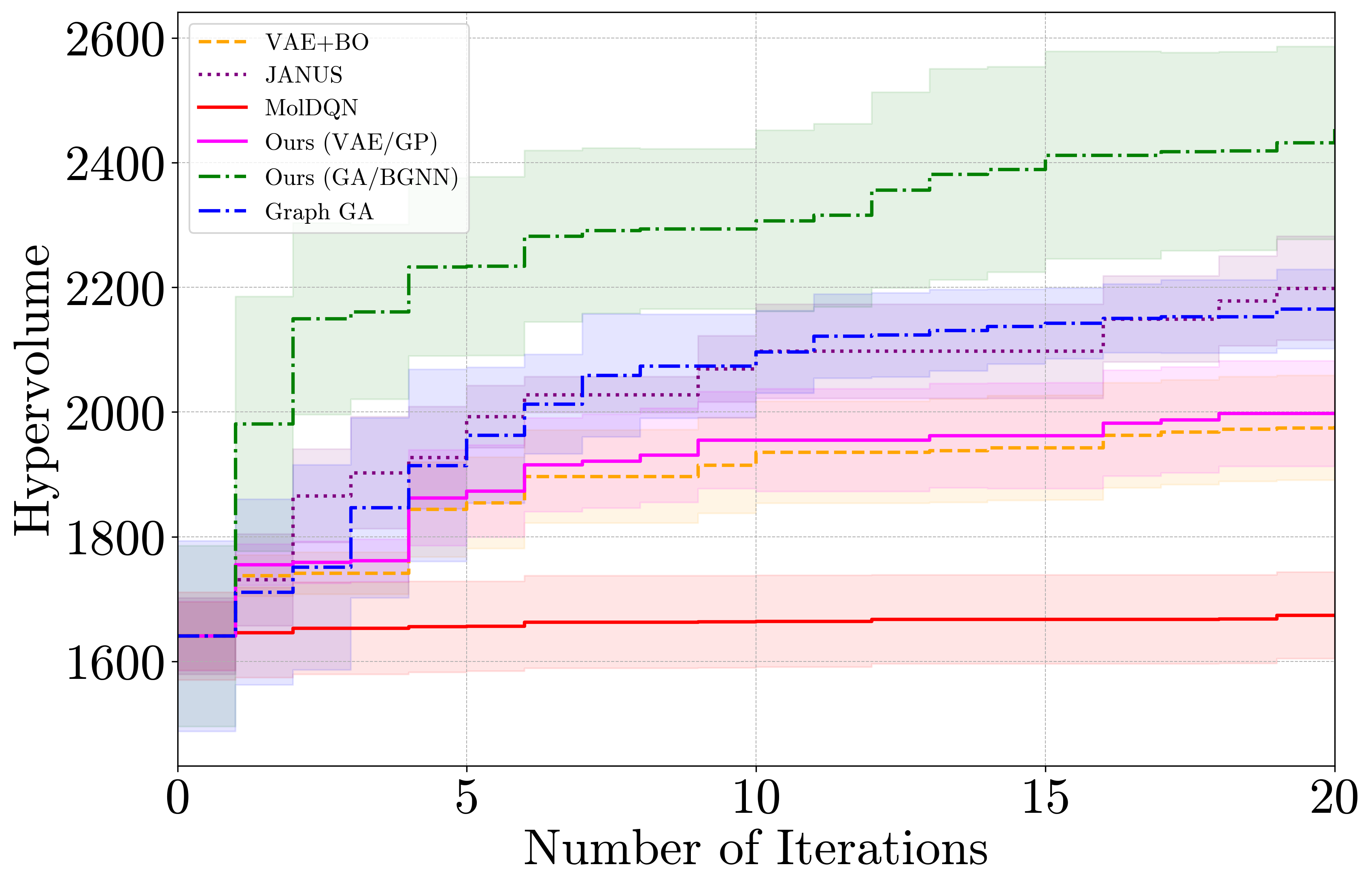}
\caption{Pareto front hypervolume progression over 20 iterations on the logP–TPSA benchmark. Each curve shows the average hypervolume achieved by a method (batch size $q = 50$), with shaded bands denoting 95\% confidence intervals. Our method (green) achieves the highest hypervolume at each step.}
\label{fig:MOO-drug-design-results}
\end{figure}

Figure~\ref{fig:MOO-drug-design-results} shows the evolution of hypervolume over 20 optimization iterations. Hypervolume is measured with respect to a fixed reference point, defined as the nadir of the initial sample set. Each curve reports the mean over five trials; shaded regions denote 95\% confidence intervals.
Our method consistently achieves the highest hypervolume across all iterations, expanding the Pareto front more rapidly and comprehensively than the baselines (see also Table~\ref{tab:MOO-drug-design-final-results}). Graph-GA performs similarly to JANUS, and both clearly outperform MolDQN, VAE+BO, and Ours (VAE/GP). These results underscore the importance of the generator: overly greedy or narrow strategies can hinder effective trade-off exploration in MOO settings.

\begin{figure}[tb]
\centering
\includegraphics[width=0.8\textwidth]{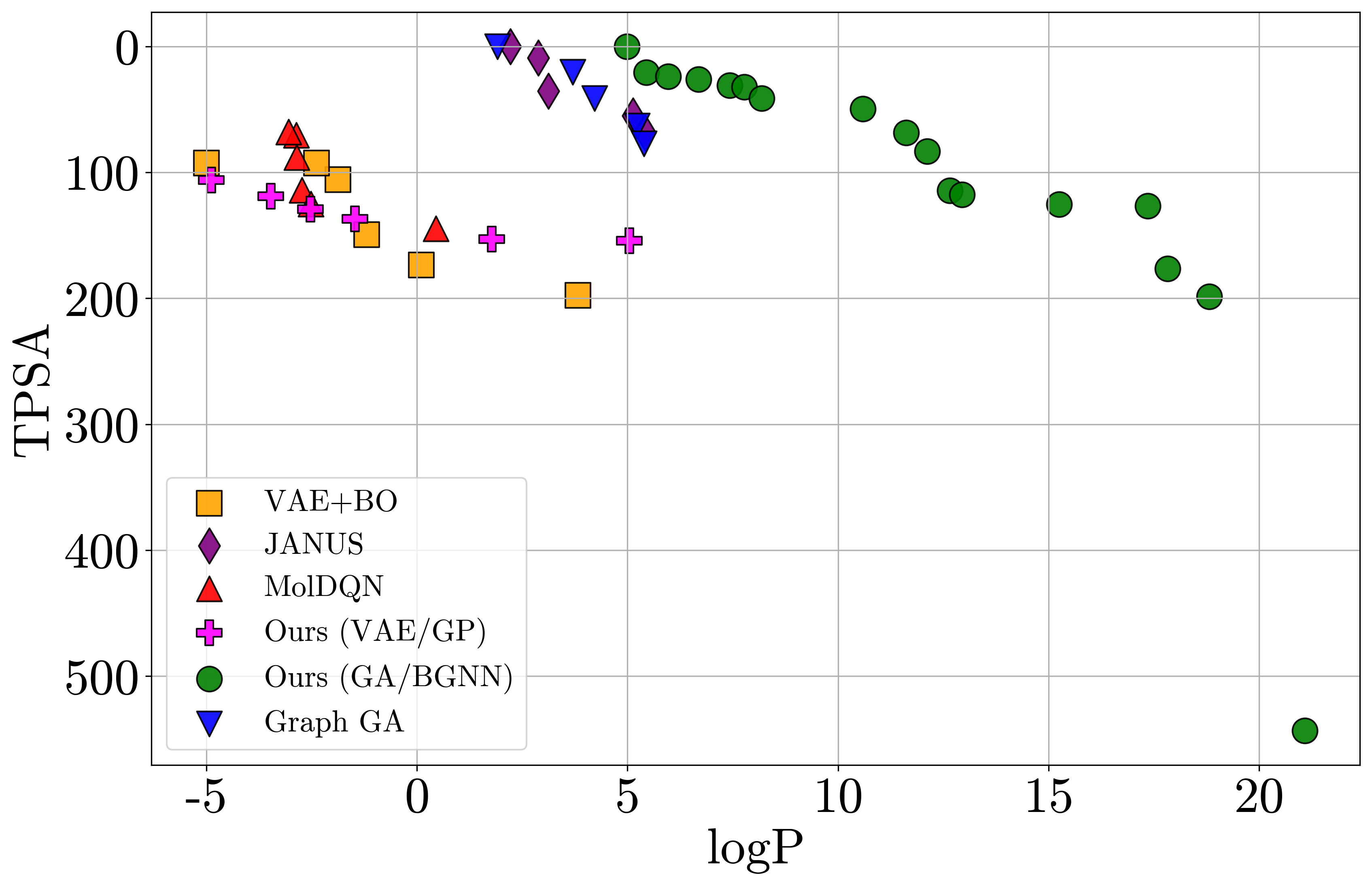}
\caption{Final Pareto fronts for each method on the logP–TPSA benchmark (best trial per method). Each point is a non-dominated molecule sampled during optimization. Our method (green) identifies a broader frontier, particularly at high-logP / low-TPSA trade-offs.}
\label{fig:MOO-drug-design-Pareto-results}
\end{figure}

Figure~\ref{fig:MOO-drug-design-Pareto-results} visualizes the final Pareto fronts (best trial per method). Our method uncovers a substantially broader frontier, including molecules with logP $>20$, well beyond the reach of other approaches. In contrast, JANUS rarely finds candidates with logP $>5$, reflecting its conservative sampling. This demonstrates our method's ability to identify both balanced and extreme trade-offs by explicitly targeting frontier expansion.

\begin{figure}[tb]
\centering
\includegraphics[width=0.7\textwidth]{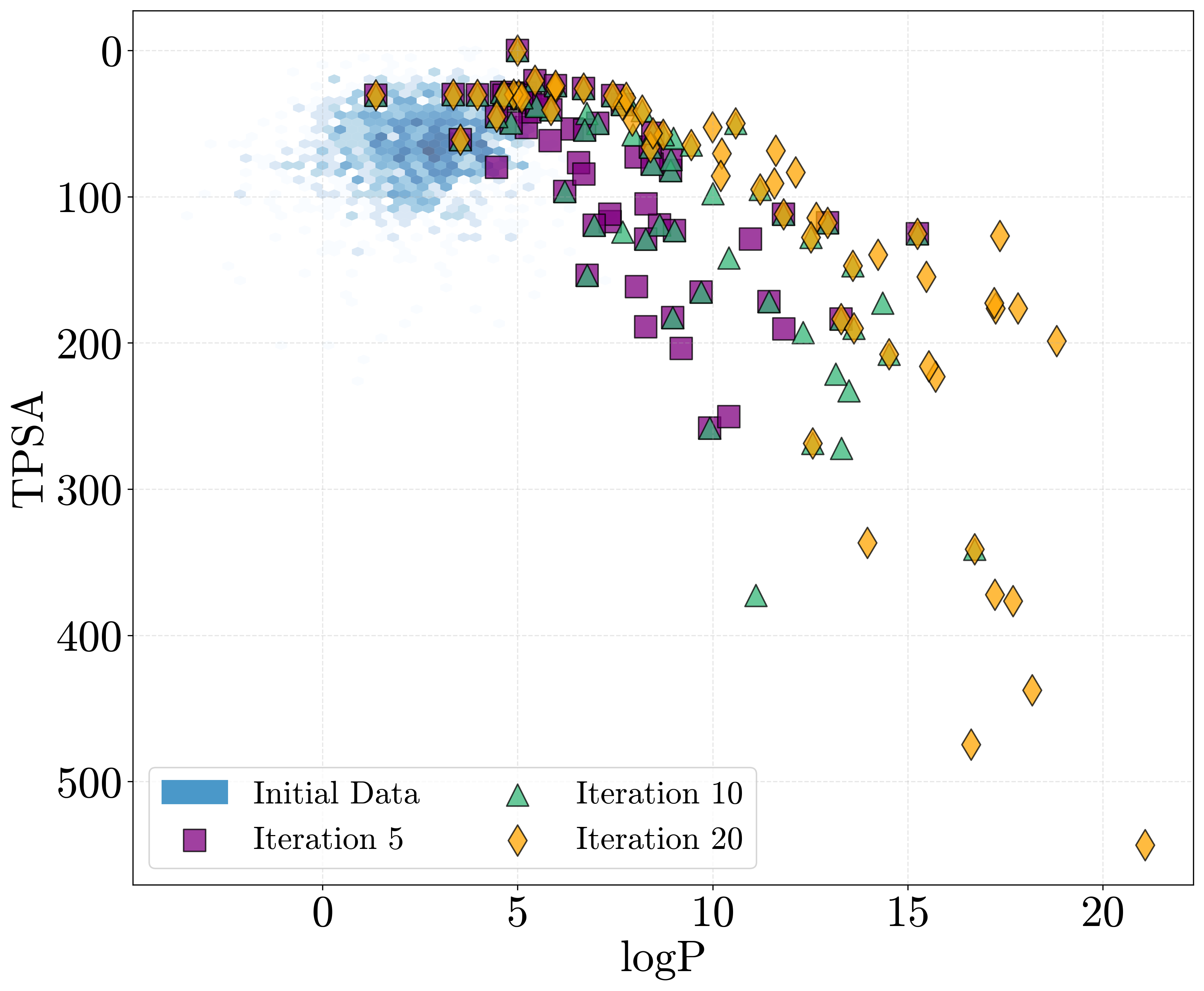}
\caption{Candidate batches selected by our method (GA/BGNN) at iterations 5 (purple squares), 10 (green triangles), and 20 (orange diamonds), overlaid on a hexbin density of the initial training data. Our method progressively expands outward from the training distribution while maintaining coverage of diverse trade-off regions.}
\label{fig:MOO-drug-design-iteration}
\end{figure}

To further illustrate this behavior, Figure~\ref{fig:MOO-drug-design-iteration} plots the evolution of our batch selections across three iterations (5, 10, and 20), overlaid on the initial data distribution. The selected molecules progressively expand into unexplored regions, while maintaining diversity across the trade-off surface. This highlights how our acquisition strategy encourages both exploration and broad Pareto coverage.

\setlength{\tabcolsep}{8pt}
\begin{table}[tb]
    \caption{Final hypervolume values across 5 independent optimization runs for each method on the multi-objective (logP-TPSA) benchmark problem.}
    \centering
    \begin{tabular}{ccccccc}
    \hline
    Run & VAE+BO  & JANUS  & Graph-GA  & MolDQN  & \specialcell{Ours \\ (VAE/GP)} & \specialcell{Ours \\ (GA/BGNN)} \\\hline
    1   & 1796.12 & 2240.83 & 2198.67 & 1586.77 & 1893.88       & \underline{2349.80}        \\
    2   & 2094.25 & 2207.73 & 2270.73 & 1720.63 & 2109.19       & \underline{2550.96}        \\
    3   & 1795.88 & 2115.02 & 2086.60 & 1554.53 & 1863.34       & \underline{2282.03}        \\
    4   & 1934.12 & 2133.12 & 2137.94 & 1743.25 & 1998.18       & \underline{2427.67}        \\
    5   & 2089.06 & 2302.93 & 2132.93 & 1784.74 & 2112.75       & \underline{2589.04}        \\\hline
    \end{tabular}
    \label{tab:MOO-drug-design-final-results}
\end{table}

\subsubsection{Impact of qPMHI acquisition function}

We now isolate the effect of our proposed qPMHI acquisition function. To do so, we fix all other components: the surrogate is a BGNN trained on 500 labeled molecules randomly drawn from ZINC-250k, and the candidate pool consists of a pre-fixed 20,000 generated molecules based on random fragment combinations over ZINC-250k. We compare qPMHI to three alternatives: (i) qEHVI, (ii) qPOTS~\cite{renganathan25a} (a multi-objective Thompson sampling method), and (iii) Sobol sampling. All acquisitions are implemented using \texttt{BoTorch}~\cite{balandat2020botorch}, following current best practices for large-batch optimization.

\begin{figure}[tb]
\centering
\includegraphics[width=1.0\textwidth]{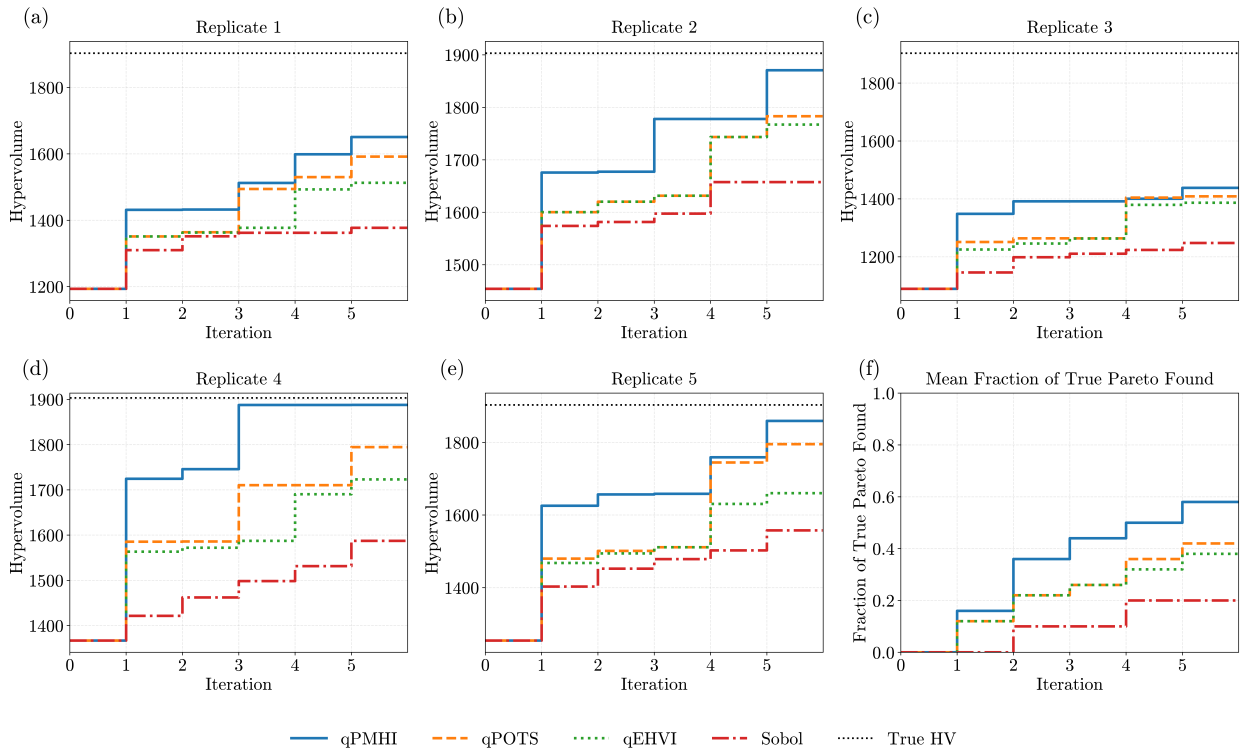}
\caption{Comparison of four acquisition functions (qPMHI, qEHVI, qPOTS, and Sobol) on the logP-TPSA benchmark using a fixed candidate pool. Subplots (a)--(e) show hypervolume over iterations for five replicates (batch size $q = 100$). Subplot (f) reports the mean fraction of true Pareto-optimal candidates recovered at each iteration. qPMHI consistently dominates in both metrics.}
\label{fig:acq-comparison}
\end{figure}

Figure~\ref{fig:acq-comparison} shows the hypervolume progression over 20 iterations (batch size $q = 100$), along with the mean fraction of true Pareto points recovered. Across all trials, qPMHI outperforms the other methods across all seeds and iterations -- achieving higher hypervolume and recovering a larger fraction of the global Pareto front. These results confirm the sampling efficiency of our new acquisition, particularly in large-batch, discrete MOO settings.

\subsection{Design of organic electrode materials}

\subsubsection{Problem description}

Quinone derivatives have emerged as promising redox-active materials for next-generation aqueous battery cathodes due to their high theoretical capacities, molecular tunability, and potential for sustainable synthesis \cite{bitenc2024organic}. Aqueous redox flow batteries (RFBs) employing such materials are particularly attractive for grid-scale energy storage, offering intrinsic safety, low cost, and independent scaling of energy and power capacity \cite{li2020material, brushett2020lifetime, hou2024towards}. However, long-term stability remains a key challenge, as molecular degradation and active species crossover can severely limit capacity retention. Identifying stable, high-voltage redox-active molecules with low aqueous solubility is therefore critical to advancing practical organic RFB technologies.

To ground our study in realistic chemistry, we leverage the large-scale dataset generated by Tabor et al. \cite{tabor2019mapping}, which contains two-electron redox potentials for over 130,000 small organic molecules (estimated via density functional theory (DFT)), including a broad set of quinones. We use this full dataset to train a BGNN surrogate to predict the redox potential $E_\mathrm{red}$ directly from molecular structure, enabling fast evaluation of new, previously unseen candidates. Aqueous solubility $S$ is predicted using FastSolv \cite{attia2025solubility}, a recently proposed ML model that provides temperature-dependent solubility estimates across a wide solvent range.

Importantly, while our property predictors are trained on existing databases, the molecules generated and evaluated during optimization are \textit{entirely novel} and lie outside the training set. Our generator is designed to construct new, synthetically plausible molecules by modifying and recombining fragments, not to sample directly from the original database (as in traditional screening). This capability is central to our framework and allows us to search large, constrained design spaces where surrogate model guidance is essential.

Because ground truth values for $E_\mathrm{red}$ and $S$ are unavailable for newly generated molecules, we rely on surrogate model predictions during optimization. The overall quality of the proposed designs is therefore limited by the accuracy and uncertainty of these models, a practical constraint that mirrors real-world discovery settings. Although our method is capable of generating high-performing candidates, our primary contribution is methodological -- we do not claim to have discovered superior molecules immediately ready for experimental testing. The DFT-calculated redox potentials in the training data are known to be approximate \cite{tabor2019mapping}, and as such, downstream conclusions should be interpreted accordingly. Our goal is to show the proposed framework can accelerate molecular design campaigns and improve search efficiency in realistic chemical spaces. This is a common strategy for benchmarking in the generative molecular optimization literature (see, e.g., Experiment II in Griffiths et al. \cite{griffiths2020constrained}). 

\subsubsection{Baseline methods}

For this case study, we compare our method to two relevant baselines: \textbf{JANUS} \cite{nigam2022parallel} and \textbf{FASMIFRA} \cite{berenger2021molecular}. JANUS was a top performer in earlier benchmarks and serves as a strong, optimization-based baseline. In contrast, FASMIFRA is a property-agnostic fragment recombination method that randomly assembles new molecules based on learned fragment distributions. This provides a naive structural baseline, useful for quantifying the impact of property-guided search.
Other methods were excluded due to their difficulty in handling structural and synthetic constraints specific to this domain. We restrict the search space to molecules containing a para-, ortho-, or anthra-quinone core, consistent with design motifs used in aqueous RFB cathodes and well-represented in the training data. Additional filters are imposed to ensure chemical validity and synthetic plausibility (see Section~\ref{subsubsec:implementation-oem}).

\subsubsection{Implementation details and setup}
\label{subsubsec:implementation-oem}

We adopt the general framework from Section~\ref{subsec:multi-objective}, but adapt it to reflect the increased realism of this domain. The initial dataset $\mathcal{D}_0$ consists of 20,000 molecules sampled from the Tabor et al. library, covering a broad spectrum of quinone derivatives. Optimization is conducted for $T = 30$ iterations.
Our GA generator is modified to enforce domain constraints: all molecules contain a valid quinone core, and genetic operations are restricted to specified attachment points. We further filter molecules to have SAScores~\cite{ertl2009estimation} below 7 and no more than six ring structures. These constraints ensure that generated molecules are chemically reasonable and remain within a space of likely synthesizable structures.
These choices demonstrate the flexibility of our framework: we show that constrained, chemically meaningful molecule generation can be achieved without requiring complex generative models, and that domain knowledge can be directly encoded into the generation pipeline. 

JANUS is applied as in the previous section, optimizing an equally weighted scalarization of $E_\mathrm{red}$ and $\log S$ based on min-max normalized training data. FASMIFRA is executed with its default settings, generating a batch of candidates at each iteration and randomly selecting the first $q = 50$ that meet all validity and constraint checks.

\subsubsection{Surrogate model performance}

\begin{figure}[h!]
\centering
\includegraphics[width=0.98\textwidth]{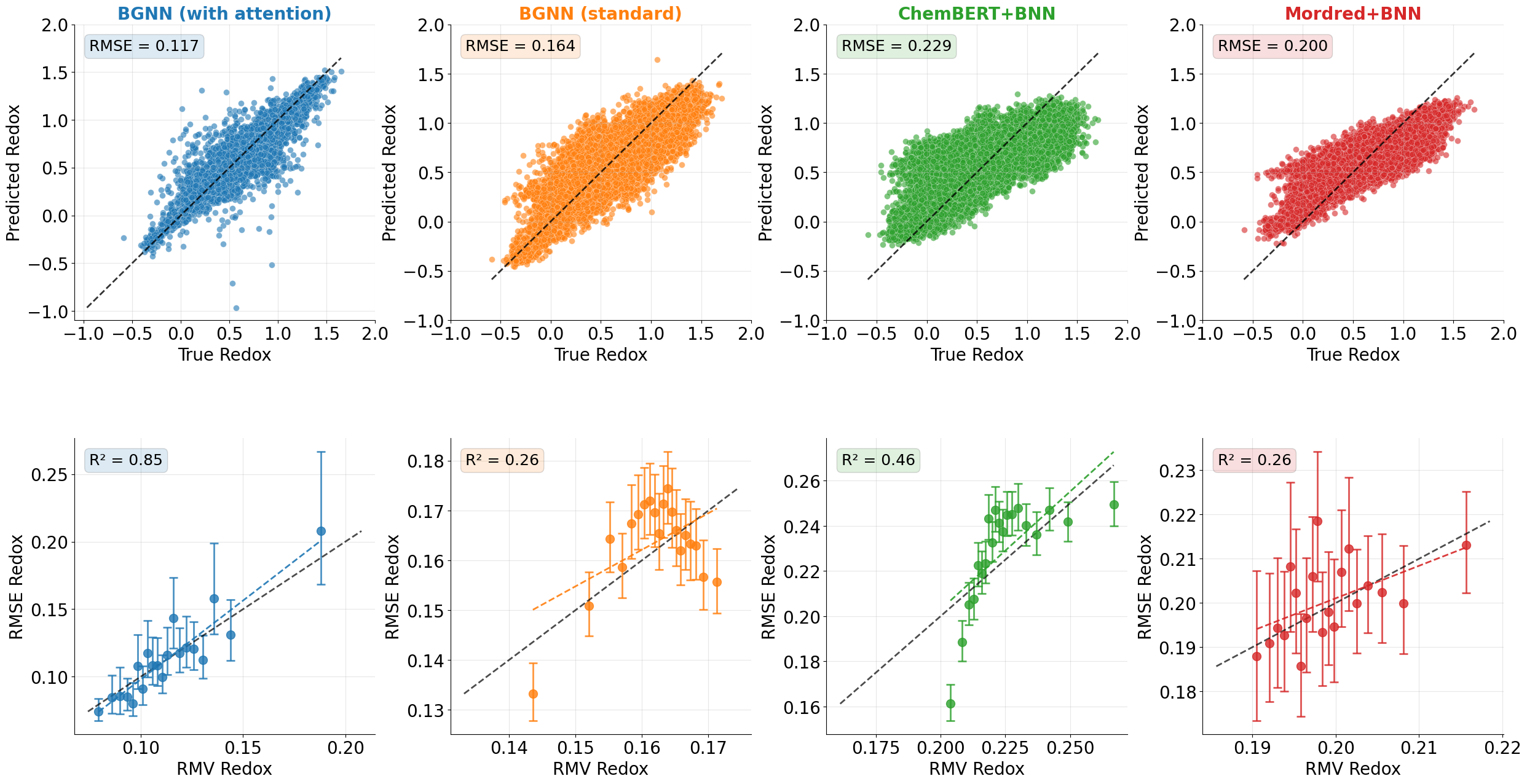}
\includegraphics[width=0.98\textwidth]{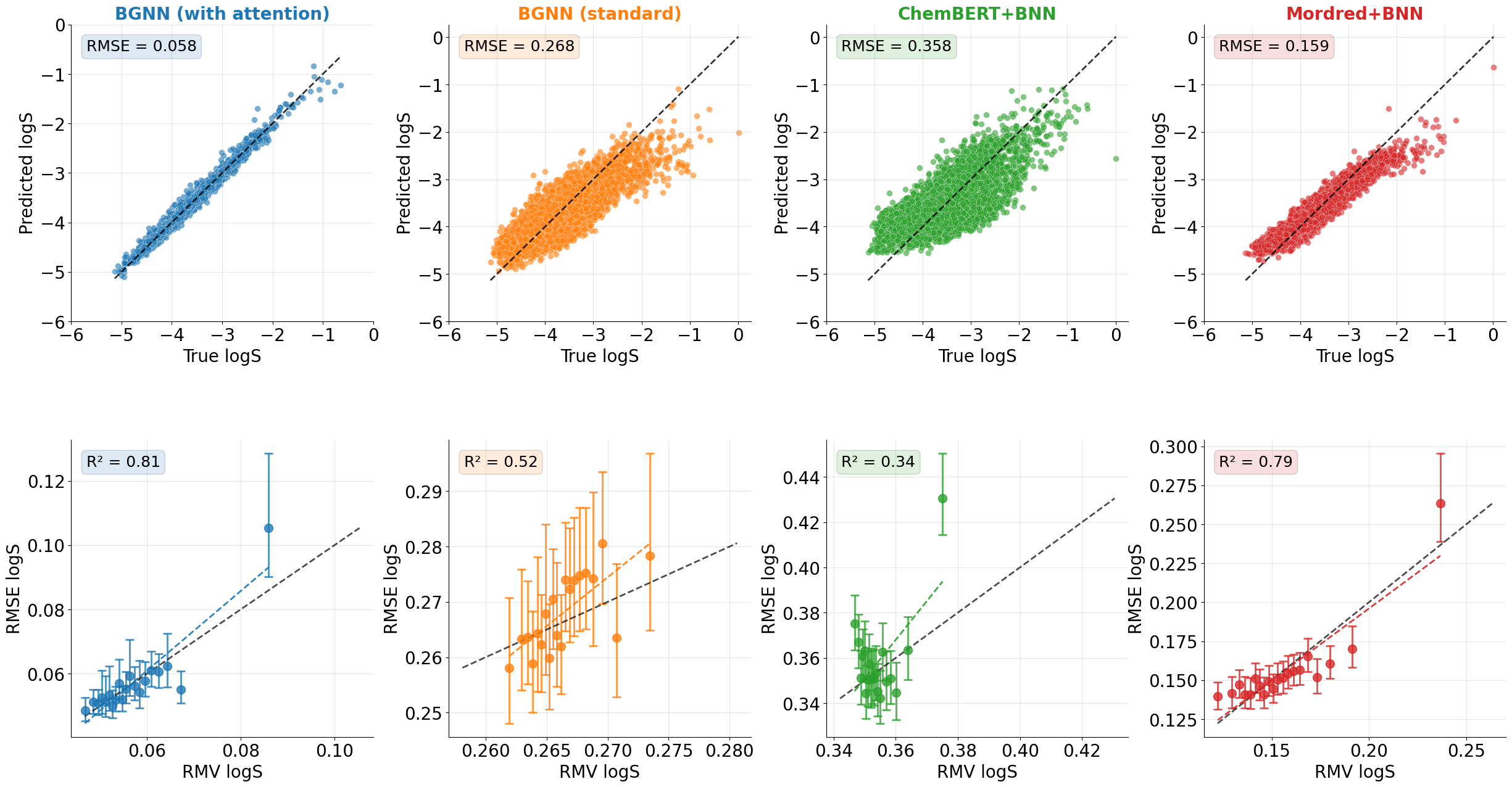}
\caption{Predictive performance of the probabilistic surrogate models (trained on the initial 20,000 dataset) on the redox potential (top two rows) and aqueous solubility (logS; bottom two rows) prediction tasks for the OEM design case study. For each property, the top row shows parity plots on a held-out test set of 15,000 molecules; the bottom row shows the root mean squared error (RMSE) versus root mean variance (RMV) computed over bins sorted by predicted uncertainty \cite{rasmussen2023uncertain}. An ideal model should exhibit a linear one-to-one relationship (dashed line), indicating well-calibrated predictive uncertainties. Our proposed model (BGNN with attention, leftmost) consistently achieves the lowest RMSE and best uncertainty calibration ($R^2 = 0.85$ for redox, $R^2 = 0.81$ for logS), outperforming all baselines.}
\label{fig:surrogate-results}
\end{figure}

We assess surrogate model performance in this realistic setting using four modeling approaches. Our proposed model, a BGNN with attention over gated recurrent units (GRUs), is compared to an ablated BGNN without attention and to two Bayesian multilayer perceptrons (MLPs) using either Mordred descriptors \cite{Mordred} or ChemBERTa embeddings \cite{chithrananda2020chemberta}.
Figure~\ref{fig:surrogate-results} summarizes the results. The proposed BGNN with attention achieves the lowest RMSE and the most accurate uncertainty calibration (using the suggested metric from Rasmussen et al.~\cite{rasmussen2023uncertain}) for both $E_\text{red}$ and $\log S$ on a held-out test set of 15,000 molecules randomly sampled from the Tabor et al. \cite{tabor2019mapping} dataset. These results indicate the model’s strong ability to generalize in a domain-constrained setting and support its use in guiding search. 

\subsubsection{Pareto optimization results}

\begin{figure}[tb]
\centering
\includegraphics[width=1.0\textwidth]{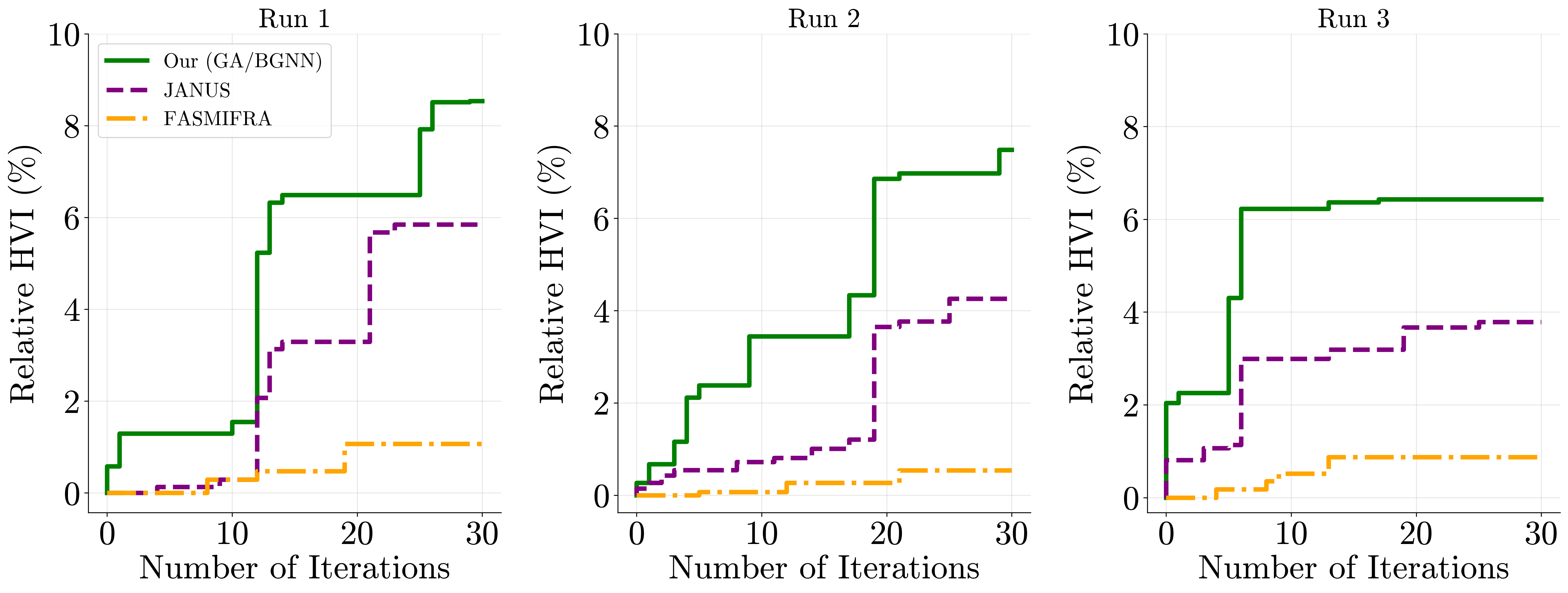}
\caption{Relative hypervolume improvement (HVI) versus number of optimization iterations for the organic electrode material (OEM) design case study. Each panel corresponds to one of three independent optimization runs. Curves show the relative gain in Pareto hypervolume with respect to the initial training set. Our proposed method (GA/BGNN; solid green) consistently achieves greater improvements across all trials compared to JANUS (dashed purple) and the property-unaware baseline FASMIFRA (dot-dashed orange), demonstrating its ability to expand the Pareto front even under structural and synthetic constraints.}
\label{fig:hvi-oem}
\end{figure}

Figure~\ref{fig:hvi-oem} shows the relative HVI for all three optimization trials, defined as $( \text{HV}( \mathcal{P}_t; \bs{r} ) - \text{HV}( \mathcal{P}_0; \bs{r} ) ) / \text{HV}( \mathcal{P}_0; \bs{r} )$, where $\mathcal{P}_t$ is the Pareto set at iteration $t$. Our method (GA/BGNN) achieves the greatest final HVI in each trial, demonstrating robust improvements over both JANUS and FASMIFRA despite the constrained and chemically challenging search space. Final HVI values and relative gains are listed in Table~\ref{tab:OEM-hypervolume-results}.
The relative improvements are smaller than in the earlier logP-TPSA benchmark, which is expected: (i) the starting dataset is larger and already contains high-quality designs, and (ii) the domain constraints significantly restrict exploration. Nonetheless, our approach consistently expands the initial Pareto front, validating its ability to operate under realistic chemical constraints.

\setlength{\tabcolsep}{8pt}
\begin{table}[tb]
    \caption{Final hypervolume achieved by each method on the OEM design case study across 3 independent optimization runs. Values in parentheses indicate the percent improvement over the hypervolume of the Pareto front computed from the initial data.}
    \centering
    \begin{tabular}{cccc}
    \hline
    Run & Ours (GA/BGNN) & JANUS & FASMIFRA \\
    \hline
    1 & \underline{18.15} (\underline{8.49}\%) & 17.71 (5.86\%) & 16.91 (1.08\%) \\
    2 & \underline{16.01} (\underline{7.52}\%) & 15.52 (4.23\%) & 14.97 (0.54\%) \\
    3 & \underline{18.12} (\underline{6.40}\%) & 17.68 (3.82\%) & 17.18 (0.88\%) \\
    \hline
    \end{tabular}
    \label{tab:OEM-hypervolume-results}
\end{table}

\begin{figure}[tb]
\centering
\includegraphics[width=0.95\textwidth]{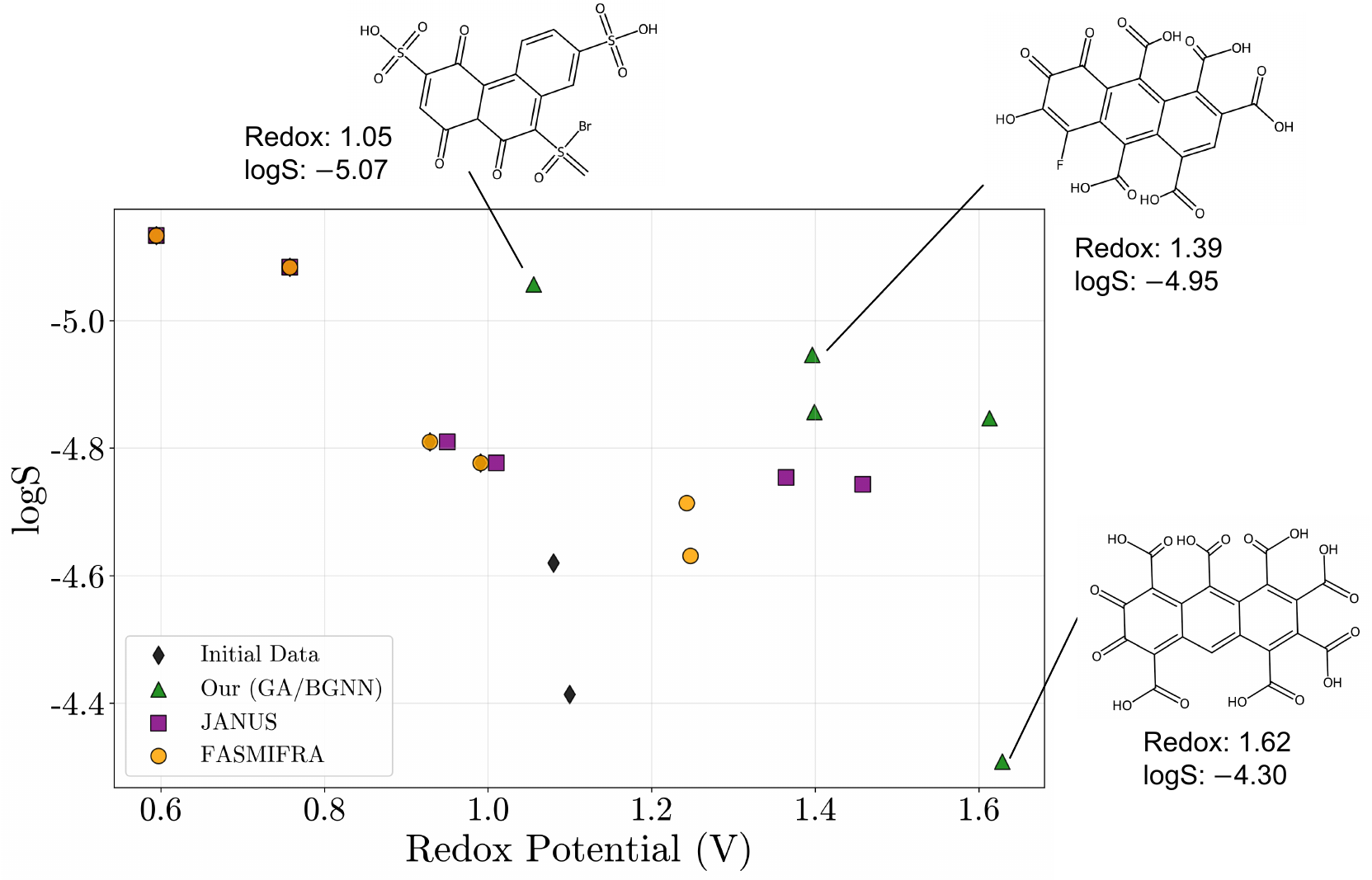}
\caption{Final Pareto fronts for the organic electrode material (OEM) design case study, taken from the highest-performing run for each method (Run 1; Table \ref{tab:OEM-hypervolume-results}). Each point corresponds to a non-dominated molecule at the end of the optimization campaign for each method. We also show the Pareto front constructed from the starting data with black diamonds. The goal is to maximize redox potential (x-axis) and minimize aqueous solubility (logS; y-axis with inverted scale), with optimal trade-offs lying toward the upper right. Our method (GA/BGNN; green) identifies the broadest front, uncovering candidates that push into underexplored regions of outcome space. Some representative molecules discovered by our approach are annotated, highlighting substitution patterns that enabled meaningful front expansion under structural and synthetic constraints.}
\label{fig:final-pareto-oem}
\end{figure}

Figure~\ref{fig:final-pareto-oem} visualizes the final Pareto fronts and representative molecules from the best-performing trial. Our method discovers new candidates that occupy previously unexplored regions of the redox-solubility trade-off space. Although we do not perform DFT or experimental validation, we note that two of the three structures depicted in Figure \ref{fig:final-pareto-oem} were identified as likely synthesizable by a synthetic chemist (the molecules with redox 1.39 and 1.62, though the synthetic routes remain uncertain). This highlights a broader challenge in molecular design: even when molecules appear chemically plausible, synthesis remains a major bottleneck due to a lack of predictive synthesis planning tools. Addressing this challenge lies outside the scope of this study, but is an important open problem for the community.

\subsection{Limitations and future work}

This work introduces a modular ``generate-then-optimize'' workflow and a new acquisition function, qPMHI, designed to improve candidate selection in multi-objective molecular design. While our primary goal was to establish the efficacy of this methodological framework, several limitations and directions for future research merit further discussion.

Stage 1 (the generator) fundamentally shapes what the optimizer can explore. Our framework does not require a large starting database -- only a generator and a small set of labeled molecules to train a surrogate. In practice, generators benefit from seeding (e.g., pretraining on generic corpora, initializing with domain scaffolds, or sampling via robust encodings like SELFIES with filters). However, generators can drift toward syntactically valid but unrealistic regions, particularly in unconstrained \textit{de novo} discovery settings. We addressed this through pragmatic filters in our case studies (e.g., attachment restrictions around quinone cores), but stronger feasibility constraints (such as synthetic accessibility metrics, retrosynthetic route planning, and stability checks) can be integrated directly into Stage 1 or treated as additional objectives. As tools like, e.g., SPARROW \cite{fromer2024algorithmic} mature, combining molecular and route design in a unified optimization loop becomes increasingly feasible.

Stage 2 depends on a probabilistic surrogate to score large batches of proposed molecules. While we demonstrated that different surrogates (e.g., BGNNs, GPs) can be used effectively, there is no single best model across properties, molecular representations, or data regimes. Improved strategies for model selection, calibration, and ensembling, potentially incorporating multi-fidelity or cost-aware formulations, are important extensions. Additionally, because our method is oracle-agnostic, any biases or noise in the underlying property estimator (e.g., DFT errors, surrogate mismatch, experimental variability) will propagate to the selection process. We did not re-evaluate proposed molecules at higher fidelity in this work; such validation is a natural next step when targeting a specific application.

Our implementation prioritizes proof-of-concept rather than computational efficiency. Per-iteration wall-clock times (Section~S2.3 of Supporting Information) are small relative to expensive simulations or experiments, but our code is not optimized and does not leverage full parallelization. 
More broadly, a time-aware formulation that allocates resources between candidate generation, surrogate training, and acquisition maximization would increase practical utility, especially in settings with fast or heterogeneous oracles.

Finally, the generate–then–optimize paradigm generalizes beyond small-molecule design. Because it operates over graph- or sequence-based representations and cleanly separates generation/proposal from selection, the same workflow could apply to polymers, inorganic materials, reaction networks, or even string-based process synthesis. These broader domains present new challenges in representation, feasibility, and scoring, but they also highlight the flexibility of the proposed approach and its potential to serve as a foundation for scalable design pipelines across chemistry, materials science, and beyond.

\section{Conclusions}
\label{sec:conclusions}

In this work, we introduce a modular framework for \textit{de novo} multi-objective molecular design that decouples candidate generation from selection. This ``generate-then-optimize'' structure supports diverse generative models and probabilistic surrogates while enabling principled selection via a new batch acquisition function, qPMHI. By ranking candidates according to their probability of delivering the maximum hypervolume improvement, qPMHI permits exact batch selection through simple sorting, avoiding combinatorial search.

Across two benchmark problems (one focused more on drug-like molecule discovery and the other on design of organic electrode materials), our framework consistently identified higher-quality Pareto fronts than strong baseline methods. Furthermore, we consistently observe that pairing qPMHI with any generator results in performance improvements, highlighting the benefits of decoupled generation and selection. The modularity of the proposed framework also ensures compatibility with emerging models, allowing users to straightforwardly incorporate improved generators or surrogates without re-architecting the pipeline.

To our knowledge, this is the first generative multi-objective BO approach that naturally supports high-throughput batch selection, making it compatible with parallel experimental platforms (e.g., multi-well screening arrays) and high-performance computing environments for large-scale simulation campaigns. Together, the presented results suggest that separating large-scale proposal from uncertainty-aware, batchwise selection is a powerful organizing principle for generative optimization in chemistry and beyond. 

\begin{acknowledgement}
The authors gratefully acknowledge support from the National Science Foundation (NSF) under Grants \#2237616 and \#2515046. The authors also thank Professor Shiyu Zhang (The Ohio State University) for insightful feedback and discussions related to the synthesizability of the molecules proposed by our method in the organic electrode material case study. 
\end{acknowledgement}

% \clearpage

\begin{suppinfo}
The Supporting Information is provided after the references. 
\begin{itemize}
    \item Description of the major classes of generative methods (Section S1), implementation details for our proposed method and the other benchmarks including a computational time comparison (Section S2), additional results for single-objective problems (Section S3), and a detailed comparison of our method to REINVENT (Section S4) (PDF).  
\end{itemize}
\end{suppinfo}

\bibliography{references}

\clearpage



\section*{Supplementary material}

# Supplementary Information for Generative Multi-Objective Bayesian Optimization with Scalable Batch Evaluations for Sample-Efficient *De Novo* Molecular Design

Madhav R. Muthyala[a,b], Farshud Sorourifar[b], Tianhong Tan[a,b], You Peng[c], and Joel A. Paulson[a,b]

[a]University of Wisconsin–Madison, Department of Chemical and Biological Engineering, Madison, MI, 53706, USA
[b]The Ohio State University, Department of Chemical and Biomolecular Engineering, Columbus, OH, 43210, USA
[c]Chemometrics, AI and Statistics, Technical Expertise and Support, The Dow Chemical Company, Lake Jackson, Texas 77566, USA

October 22, 2025

# Contents

# S1 Major Classes of Generative Models

Stage 1 in our framework is intentionally *generator-agnostic*: any method capable of producing a large pool of valid molecules can plug in, and Stage 2 handles the multi-objective down-selection. Below we review the main classes of molecular generators, how they function, and the trade-offs they bring in terms of sample efficiency, validity, controllability, and coupling to surrogate models. Our aim is not to prescribe a winner but to clarify when each approach is most appropriate and what limitations often motivate our decoupled "generate-then-optimize" design.

## S1.1 Variational autoencoders (VAEs)

VAEs embed discrete molecule representations into a continuous latent space, then decode latent vectors back into molecular structures [1]. In training, the encoder and decoder are jointly optimized via a reconstruction loss plus a Kullback-Leibler (KL) divergence regularizer; property conditioning can be added via auxiliary predictors or conditional VAE variants [2]. Once trained, new molecules are generated by sampling latent vectors and decoding them.

VAEs offer smooth latent interpolations and support gradient-based navigation in latent space, which is appealing when optimizing a single (or scalarized) objective. But in practice, the continuous latent space may not align well with discrete chemical structure, making surrogate modeling and uncertainty quantification difficult, especially in multi-objective settings. Decoding may yield invalid or redundant molecules, and batch acquisition in latent space often must rely on heuristics. Some works tackle this via local BO in latent space or trust-region strategies [3]. In our pipeline, VAEs are best deployed as fast proposers of diverse candidates; Stage 2 then handles the rigorous multi-objective ranking without requiring joint training of encoder, decoder, and surrogate. Empirically, VAE-based generators often underperform compared to alternatives in multi-objective molecule discovery [4], but they remain useful baselines and fallback options.

## S1.2 Autoregressive models

Autoregressive models generate molecules one token or one graph action at a time. For example, when using SMILES/SELFIES strings, each token is predicted conditioned on previous tokens. Alternatively, graph-based autoregressive models decide on atoms, bonds, or edges step-by-step. Training typically uses maximum likelihood over known molecules; conditioning is introduced via fine-tuning or conditioning prompts.

These models are strong at validity and language-like sequence modeling. They can easily incorporate domain priors via prompts or context, but they are sensitive to representation choices (e.g. SMILES canonicalization), suffer from exposure bias (errors propagate during generation), and often stay close to training distributions without stronger exploration forces. In tightly coupled schemes, autoregressive models may over-optimize toward surrogate objectives, reducing diversity.

In our framework, autoregressive models act as scalable, high-validity proposers: they can rapidly emit large pools of structurally plausible candidates biased toward relevant chemical spaces. Stage 2 then intervenes to restore principled trade-offs by selecting the subset that maximizes expected Pareto improvement. This decoupling helps avoid overfitting generation to the surrogate.

## S1.3 Diffusion and flow models

Diffusion models generate molecules by progressively denoising a random latent through a learned transition, whereas normalizing flows map molecules and latent vectors via invertible transformations. Both approaches support conditioning (e.g. score-based guidance in diffusion, conditional flows) and tend to produce diverse, high-quality samples, including geometry-aware variants for three-dimensional molecular structures [5]. Flow-autoregressive hybrids (e.g. GraphAF [6]) combine the benefits of flows and autoregressive sampling with validity checks during generation.

These models excel when sample quality and diversity matter most, and when geometric consistency or symmetry is important. However, training and sampling can be more compute-intensive, and coupling them tightly to property surrogates risks over-optimization. In our generate-first paradigm, diffusion/flow models

can generate a broad structural scaffold in bulk (often in unconditional or lightly guided mode), leaving Stage 2 to impose multi-objective constraints and select the most promising candidates.

## S1.4 Evolutionary methods

Evolutionary algorithms (EAs) evolve populations of molecules via mutation, crossover, and selection directly in chemical space. A prominent example is Graph-GA [7], which has been demonstrated to traverse structural space effectively (e.g., discovering novel scaffolds in logP optimization tasks). More recent work highlights how GAs remain strong baselines [8]. In multi-objective molecular optimization, variants of NSGA-II and NSGA-III have been adapted to work over molecular graphs, enforcing diversity and Pareto advancement without relying on learned models [9]

Graph-GA is attractive because it is simple, interpretable, and fairly flexible in terms of the graph representation. It naturally incorporates domain heuristics (e.g. synthesizability filters, substructure constraints) and does not require training. However, it can be evaluation-inefficient and prone to local minima if diversity is not carefully enforced. In our framework, Graph-GA can serve as an adaptive proposer: we could bias mutation/selection pressures toward molecules in regions of high surrogate uncertainty or predicted potential, then hand off the candidate set to Stage 2, which guards against premature convergence by selecting across the population in a multi-objective, uncertainty-aware fashion.

## S1.5 Reinforcement learning (RL)

RL treats molecular generation or editing as a sequential decision problem: at each step the agent chooses an action (e.g. add atom, bond, transform fragment), receives a reward based on predicted properties (or multi-objective scalarization), and updates its policy. Early molecular RL methods include REINVENT [10], which applies policy gradient to sequence-based generators (SMILES), and MolDQN [11], which frames editing actions in a Q-learning framework. More advanced methods also combine autoregressive or flow models with RL or GFlowNet-style training for better diversity and sampling properties. GFlowNets (generative flow networks) aim to sample terminal molecules in proportion to a reward distribution, thus inherently promoting diversity instead of greedily chasing the highest reward [12]. Recent variants have incorporated goal-conditioning and multi-objective extensions [13].

RL and GFlowNets offer fine-grained control, curriculum-style learning, and constraint enforcement; however, they tend to require many reward evaluations, careful reward shaping, and suffer from instability or mode collapse. In our decoupled setup, RL or GFlowNet agents can be used to enrich regions of interest (e.g., via short rollouts targeted to promising areas) while Stage 2 maintains rigorous uncertainty-driven batch selection, limiting overfitting of the surrogate.

# S2 Implementation Details

## S2.1 Benchmark methods

For the initial case studies, we compare our proposed method (described next) to four alternative methods. Each of these baselines represents a different family of generative optimization techniques. Below, we describe their implementation and how we adapt them for the multi-objective setting:

- **VAE+BO** [1]: This approach follows the latent-optimize-then-decode framework introduced by Gómez-Bombarelli et al., where molecules are embedded into a continuous latent space using a VAE, optimized with BO, and then decoded back into molecular structures. We implement the same architecture as the original paper, with a sparse GP surrogate model and the qLogEI acquisition function (for single-objective problems) or qLogEHVI (for multi-objective problems), both implemented using `BoTorch` [14]. LogEI and LogEHVI offer numerically stable approximations of Expected Improvement (EI) and Expected Hypervolume Improvement (EHVI), respectively, that are well-suited for multi-point (batch) optimization [15].

- **JANUS** [16]: JANUS is a parallel-tempered GA that integrates a neural network property predictor to guide evolutionary selection. It uses multiple interacting populations at different "temperatures"

to balance exploration and exploitation. We adopt the default configuration and hyperparameters provided in the original implementation, available at https://github.com/aspuru-guzik-group/JANUS. As JANUS is designed for single-objective optimization, we extend it to multi-objective settings by optimizing a composite scalar objective defined as a weighted sum of the normalized objectives, with weights determined based on min-max scaling of the initial dataset.

- **Graph GA** [7]: This method is a graph-based GA that builds molecules through mutations and crossovers of molecular fragments. It starts with a seed population of valid molecules and applies chemically informed transformations, using a molecular graph representation and `RDKit` sanitization to enforce valency and stability constraints. The fitness function is based on a target property and selection favors molecules with improved performance. We use the default configuration and hyperparameters described in the original paper and implementation, available at https://github.com/jensengroup/GB_GA. To adapt Graph GA to multi-objective problems, we again apply a scalarization strategy using min-max normalized objectives and fixed weights.

- **MolDQN** [11]: MolDQN is a deep reinforcement learning method based on Q-learning that incrementally edits molecular graphs to optimize desired properties. It models the problem as a Markov decision process, where each action corresponds to a chemical transformation. The policy is trained to maximize a reward function, which we take as a scalarized version of the multi-objective targets. We use the original implementation and training procedure from https://github.com/google-research/google-research/tree/master/mol_dqn, with no changes to architecture or hyperparameters. For multi-objective problems, we define the reward as a linear combination of normalized objectives using fixed weights (same as JANUS and Graph GA).

**Constraint handling**: For case studies where structural constraints are required to prevent trivial solutions (e.g., overly long SMILES strings in logP optimization), we incorporate filtering or rejection mechanisms specific to each method. In VAE+BO, we inject increasing Gaussian noise into the latent vector until a decoded molecule satisfies the constraints. For JANUS and Graph GA, we apply post-generation filtering on the candidate pool before the final batch selection to be sent to oracle evaluation. For MolDQN, we regenerate batches until a valid set of constraint-satisfying molecules is found.

The full implementation details, including hyperparameters and evaluation pipelines for all baseline methods, are available in our code repository: https://github.com/PaulsonLab/Generative_MOBO_qPMHI.

## S2.2 Our proposed method

Our approach follows a modular two-stage design framework. Stage 1 involves the generative proposal of candidate molecules, while Stage 2 performs surrogate-based batch selection using our novel acquisition function (i.e., qPMHI).

**Stage 1: Generative proposal.** We explore several generator types to highlight both the flexibility of our framework and the role of the generator in optimization performance. Our default choice is a lightweight genetic algorithm (GA) applied to the SELFIES representation of molecules, guided by the surrogate model used in Stage 2. Let $\widehat{\boldsymbol{f}} : \mathcal{X} \to \mathbb{R}^M$ and $\boldsymbol{\Sigma} : \mathcal{X} \to \mathbb{R}^{M \times M}$ denote the surrogate's predictive mean and covariance functions. Candidates are generated by maximizing a composite scoring function:

$$J_{\boldsymbol{\lambda}}(\boldsymbol{s}; \widehat{\boldsymbol{f}}, \boldsymbol{\Sigma}, \boldsymbol{g}),$$

where $\boldsymbol{s}$ is a SELFIES string, $\boldsymbol{g}$ denotes optional auxiliary metrics (e.g., synthesizability), and $\boldsymbol{\lambda}$ is a tunable parameter vector that balances the components of the score. Rather than fixing $\boldsymbol{\lambda}$, we sweep over many random instantiations (typically drawing each coefficient uniformly between 3 and 5) and aggregate the resulting unique molecules across GA runs. This ensemble-style approach encourages candidate diversity and balances exploitation (high predicted utility) with exploration (uncertain regions or auxiliary objectives). We implement the GA using the `DEAP` Python package [17], following the STONED methodology [18], which leverages the robustness of SELFIES to avoid invalid molecule generation.

In addition to the default GA, we evaluate two alternative generators. The first is a VAE-based generator, reusing the same trained model from the VAE+BO baseline. Here, we sample latent points in the

4

Table S.1: Wall-clock time per optimization iteration (mean ± std, seconds) on the logP-TPSA study. Note that these methods were not all run on the same computing cluster or optimized for speed, so results should be viewed as approximate indicators.

| Method | Time per iteration (min) |
|---|---|
| Ours[†] | $11.71 \pm 0.10$ |
| JANUS[†] | $12.63 \pm 0.57$ |
| VAE+BO[‡] | $11.95 \pm 0.52$ |
| VAE+GP[‡] | $11.75 \pm 0.05$ |
| MolDQN[‡] | $11.44 \pm 0.05$ |
| Graph-GA[§] | $2.21 \pm 0.03$ |

[†]Run on computer with 16 CPU cores, 512 GB RAM.
[‡]Run on computer with 16 CPU cores, 512 GB RAM, NVIDIA A100 (40 GB).
[§]Run on computer with 48 CPU cores, 192 GB RAM; heavily parallelized run.

neighborhood of those proposed by BO and decode them into molecules. The second is a REINVENT-style generator, which uses a recurrent neural network trained via reinforcement learning to propose molecules with favorable properties (see Section S4 for details). For all generators, if the initial candidate pool does not meet any potential desired size or constraint requirements, we re-run the generator with new random seeds until a sufficiently large, valid pool is obtained.

**Stage 2: Surrogate modeling and batch selection.** For Stage 2, we develop a custom Bayesian graph neural network (BGNN) surrogate model. Each molecule is represented as a molecular graph, parsed from its SMILES string, with atom-level node features and bond-level edge features. The encoder is a deterministic graph attention network with message passing, adapted from Ramani and Karmakar [19], which maps each graph to a 200-dimensional latent feature vector. This embedding is passed through a series of Bayesian linear layers with ReLU activations to produce property predictions.

The BGNN is trained via variational inference to minimize the Kullback-Leibler (KL) divergence between the true posterior and a tractable variational approximation. Optimization is performed using the AdamW optimizer with a learning rate of 0.001. We use the `torchbnn` package to implement the variational layers, and apply the post-hoc uncertainty calibration method proposed by Rasmussen et al. [20] to improve predictive variance estimation. For computational efficiency, we update the BGNN parameters every 5 optimization iterations, with each update using 100 training epochs (i.e., full passes through the dataset).

## S2.3 Computational time comparison

Table S.1 reports the (end-to-end) wall-clock time per optimization iteration for the multi-objective logP-TPSA case study (reported in Section 4 of the main body). Values are mean ± standard deviation (minutes), computed across iterations of all replicates. Two caveats are important for interpreting these numbers. First, due to practical constraints on software installation and cluster availability, not all baselines were executed on identical hardware; we indicate the configuration for each method in the table footnote. Second, we did not tune any code for speed. For the applications we target – where the oracle is a high-fidelity simulation or an experiment – minutes per iteration are typically negligible compared to evaluation time, which can be hours to days or longer. Nevertheless, scenarios with cheap oracles may warrant additional engineering; we view systematic runtime optimization as future work.

# S3 Single-Objective Optimization of logP

We assess the generate-then-optimize framework on a single-objective benchmark to enable direct comparison with widely used baselines that are primarily tailored for single-objective settings.

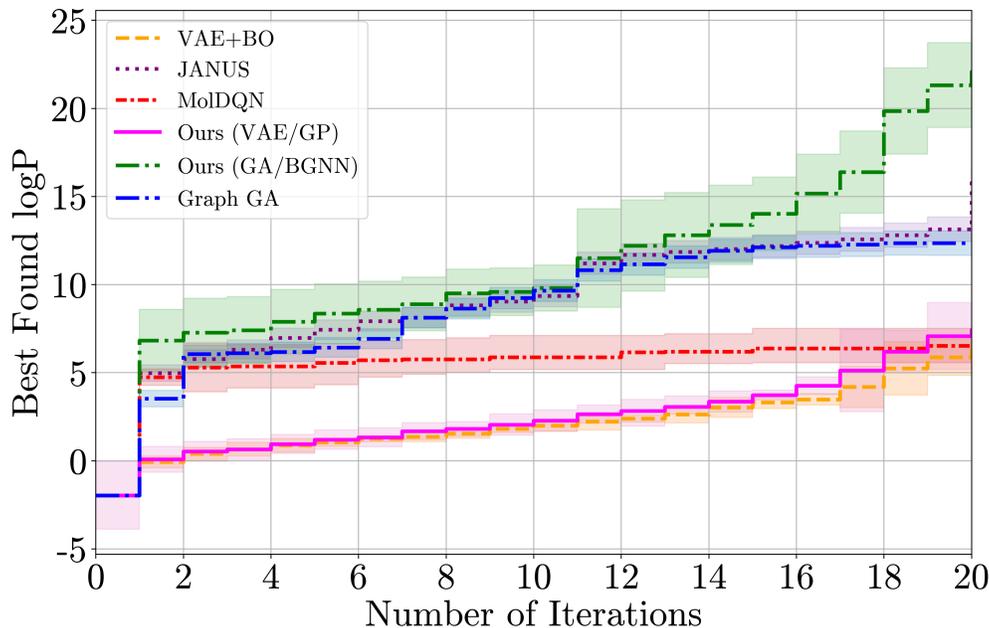

Figure S.1: Single-objective logP optimization results across 20 iterations (batch size $q = 50$). Curves show mean best-found logP across five seeds; bands denote 95% confidence intervals. Our method, *Ours (GA/BGNN)* (green), consistently outperforms all baselines.

## S3.1 Problem setup

We maximize the water–octanol partition coefficient (logP), following the Gómez-Bombarelli et al. setup [1]. Molecules are drawn from ZINC-250k [21] and logP is computed using the Crippen method [22] via `RDKit` [23]. To avoid trivial "hacks" that are known to exist for logP, we enforce two constraints throughout: SMILES length $\leq 108$ and SAScore $\leq 8$ (which is a measure of the ease of synthesizing a molecule).

All methods are implemented as in Section S2. We run $T = 20$ iterations with batch size $q = 50$. For our method, we use $L = 256$ Monte Carlo samples to compute acquisition values and set the Stage 1 pool size to $N = 5000$. The initial training set $\mathcal{D}_0$ contains 2000 randomly sampled, labeled molecules from ZINC-250k. Each method is run with five random seeds; the same five seeds are used across methods to ensure consistency.

## S3.2 Results

Figure S.1 shows the best-found logP versus iteration for six methods (means over five seeds with 95% confidence intervals). Our approach, *Ours (GA/BGNN)*, using qPO for single-objective selection, improves most rapidly and attains the highest final values (exceeding 20 on average). The next strongest methods are the two GA families: *JANUS* and the newly added *Graph-GA*, which track closely and plateau near the mid-teens by iteration 20. *MolDQN* and the two VAE-based baselines remain substantially lower. Table S.2 reports the final best-found logP for each seed. Our method is best on every replicate. As a point of reference for the Stage 2 effect, *Ours (VAE/GP)*, which retains the same VAE generator but uses our two-stage selection, improves noticeably over *VAE+BO*.

## S3.3 Discussion

Two observations stand out. First, *JANUS* and *Graph-GA* (both GA-based) perform in the same band and above MolDQN and the VAE-based baselines. *Graph-GA* averages $14.93 \pm 0.32$ across seeds, while *JANUS* averages $15.78 \pm 0.29$, consistent with the side-by-side trajectories in Figure S.1. This suggests that evolutionary search with chemically informed operators and diversity pressure is an effective strategy on

Table S.2: Final best-found logP across five independent runs (one per seed). The Graph-GA baseline was newly added and performs similarly to JANUS. The best value in each row is underlined.

| Run | VAE+BO | JANUS | Graph-GA | MolDQN | Ours (VAE/GP) | Ours (GA/BGNN) |
|-----|--------|-------|----------|--------|---------------|----------------|
| 1   | 6.21   | 16.14 | 14.66    | 6.68   | 7.21          | <u>25.26</u>   |
| 2   | 6.15   | 15.78 | 15.10    | 6.06   | 7.64          | <u>24.09</u>   |
| 3   | 6.49   | 15.41 | 15.30    | 6.53   | 7.59          | <u>16.32</u>   |
| 4   | 6.30   | 15.96 | 15.05    | 6.96   | 7.24          | <u>19.92</u>   |
| 5   | 6.44   | 15.60 | 14.54    | 6.37   | 7.50          | <u>24.82</u>   |

this benchmark. Second, our two-stage method delivers a larger and more consistent improvement: *Ours (GA/BGNN)* averages $22.08 \pm 3.86$, with fast early gains and several large jumps later in the run. The *Ours (VAE/GP)* versus *VAE+BO* comparison isolates the contribution of the selection stage: holding the VAE generator fixed while switching to our acquisition-driven, batchwise selection yields a clear boost (from $6.32 \pm 0.15$ to $7.44 \pm 0.20$ on average), underscoring the value of decoupling generation from objective-driven selection even with imperfect generators.

Overall, the single-objective results reinforce two principles that motivate the full framework: (i) the generator matters, with GA families being strong in this application and (ii) principled, acquisition-based selection confers additional gains beyond generator choice, which becomes critical as we transition to *de novo* multi-objective problems.

## S4 Comparison to REINVENT

To further understand the generalizability of our proposed two-stage generate-then-optimize framework, we conducted additional experiments comparing it against REINVENT [10], a widely used RL-based molecular generator. REINVENT has demonstrated strong performance (ranked number 1 out of the tested options) in prior benchmarks for molecular property optimization [4]. These comparisons allow us to better assess the complementarity and robustness of our proposed qPMHI-based selection strategies when paired with different generative models, as well as to highlight challenges that arise in distributional mismatch between training and target domains.

### S4.1 Multi-objective optimization of penalized logP and TPSA

REINVENT uses a recurrent neural network (RNN) over SMILES strings, trained initially on ∼1.5M canonical molecules from the ChEMBL database [24]. During optimization, it performs policy-gradient updates using an augmented likelihood:

$$\log P_{\text{aug}}(\boldsymbol{x}) = \log P_{\text{prior}}(\boldsymbol{x}) + \sigma S(\boldsymbol{x}),$$

where $S(\boldsymbol{x})$ denotes a task-specific score and the tuning parameter $\sigma$ balances exploration and exploitation. To ensure a fair comparison, we reuse the same REINVENT-trained prior and focus on a two-objective task: maximize penalized logP (standard logP minus a synthetic accessibility score and ring penalty) and minimize topological polar surface area (TPSA). Note that this is slightly different than the constrained logP studied in the main body and Section S3.

**Setup.** We warm-start both methods with 10 iterations of REINVENT using its default parameters ($\sigma = 128$, learning rate $5 \times 10^{-4}$) and a batch size $q = 128$. After this shared phase, we train a surrogate GP model using Extended Connectivity FingerPrints (ECFP) (radius 3) with a Tanimoto kernel [25, 26]. For iterations 11 onward, REINVENT continues standard RL fine-tuning, while our method applies qPMHI selection to a large oversampled pool of 5000 SMILES generated from REINVENT rollouts (100 rollouts of length 50 on the surrogate mean). This ensures both methods operate in the same molecular prior space.

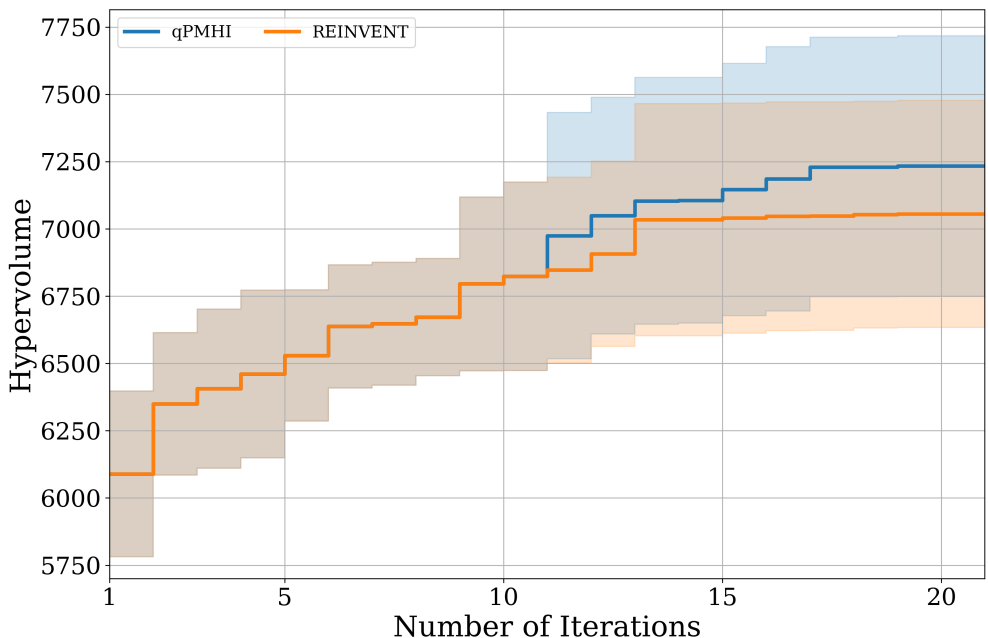

Figure S.2: Pareto front hypervolume over 20 optimization iterations on the penalized logP-TPSA benchmark. Both methods share the same REINVENT-generated prior and are identical for the first 10 iterations. From iteration 11 onward, the proposed method (qPMHI) applies surrogate-based batch selection on a large REINVENT-generated pool. Curves show the mean over 5 runs; shaded regions indicate 95% confidence intervals. Results show that augmenting REINVENT with qPMHI consistently improves performance.

**Results.** Figure S.2 presents the hypervolume of the Pareto front versus number of iterations, averaged across five independent runs. After the initial shared phase, switching to qPMHI produces a sharp increase in performance at iteration 11, and this advantage is maintained through iteration 20. Across iterations 12-20, our method consistently outperforms REINVENT by 100–200 hypervolume units, with the qPMHI confidence bands remaining elevated throughout. These results show that REINVENT, while capable of generating diverse candidates, can clearly benefit from an uncertainty-aware selection mechanism. This further reinforces the central idea behind our framework: that generator-agnostic optimization via acquisition-guided selection offers a principled and effective way to enhance performance.

### S4.2 Design of organic electrode materials

We also tested REINVENT on our more challenging multi-objective design task involving organic electrode materials (OEMs), where the goal is to jointly optimize redox potential and aqueous solubility. Unlike the other benchmark problems focused on drug-like molecules, this task involves more electrochemically diverse structures from a quinone-rich design space curated by Tabor et al. [27]. As before, we use the default REINVENT configuration and the same ChEMBL-trained prior model. However, this introduces a distributional mismatch: the ChEMBL prior is heavily biased toward small bioactive molecules, while the OEM library contains larger, more functionalized redox-active species.

**Results.** Figure S.3 plots hypervolume progression for REINVENT over 100 iterations with batch size 128. While REINVENT explores a number of molecules, its final Pareto front reaches a Pareto front hypervolume of only ∼1.34, which is substantially worse than the initial $\mathcal{D}_0$ (14.89) used in our main experiments. These results confirm that REINVENT, in its default form, fails to make meaningful progress in this application. While we expect that retraining the prior on a more relevant dataset would substantially improve performance, that is beyond the scope of this study. Nonetheless, this experiment highlights an important takeaway: the alignment between generator training distribution and target design space plays a crucial role

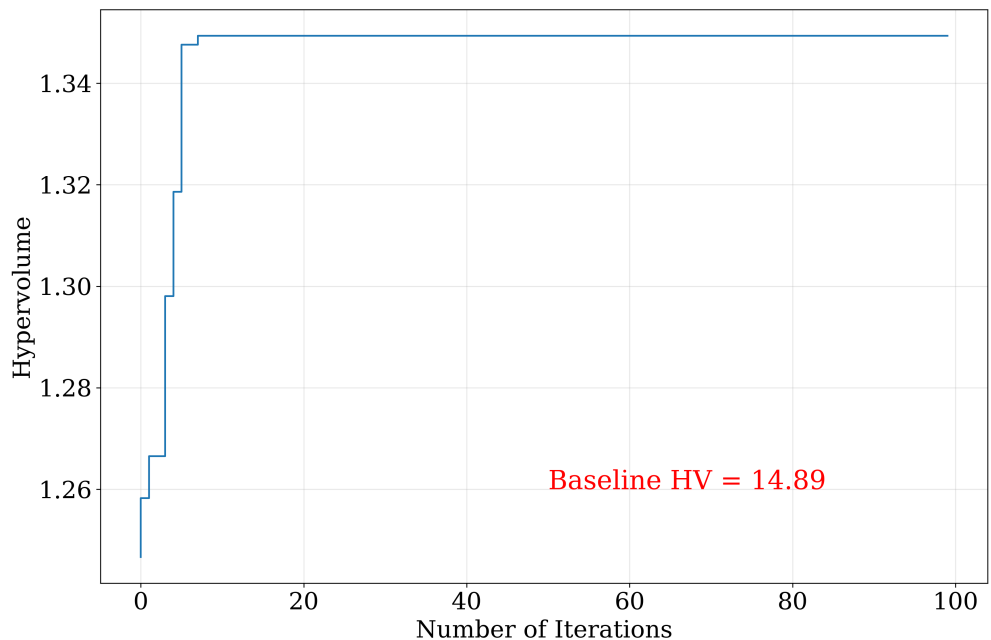

Figure S.3: Performance of REINVENT on the multi-objective organic electrode material (OEM) design task. Shown is the Pareto front hypervolume across 100 iterations (batch size 128) for a single run. The red label indicates the baseline hypervolume (14.89) of the initial set $\mathcal{D}_0$ used in the main paper. Despite running for many iterations, REINVENT fails to improve over the initial dataset due to poor alignment between its ChEMBL-trained prior and the quinone-rich OEM design space.

in downstream optimization outcomes and should be carefully considered when selecting generative models for molecular discovery.

9



\end{document}